\documentclass{article}

\usepackage{microtype}
\usepackage{graphicx}
\usepackage{subfigure}
\usepackage{booktabs} % for professional tables

\usepackage{hyperref}

\usepackage[accepted]{icml2025}

\usepackage{amsmath}
\usepackage{amssymb}
\usepackage{mathtools}
\usepackage{amsthm}

\usepackage[capitalize,noabbrev]{cleveref}

\theoremstyle{plain}
\newtheorem{theorem}{Theorem}[section]

\newtheorem{lemma}[theorem]{Lemma}

\theoremstyle{definition}
\newtheorem{definition}[theorem]{Definition}

\theoremstyle{remark}

\usepackage[textsize=tiny]{todonotes}
\usepackage[utf8]{inputenc} % allow utf-8 input
\usepackage[T1]{fontenc}    % use 8-bit T1 fonts
\usepackage{hyperref}       % hyperlinks
\usepackage{url}            % simple URL typesetting
\usepackage{booktabs}       % professional-quality tables
\usepackage{amsfonts}       % blackboard math symbols
\usepackage{nicefrac}       % compact symbols for 1/2, etc.
\usepackage{microtype}      % microtypography
\usepackage{xcolor}         % colors
\usepackage{multirow}
\usepackage{amsmath}
\usepackage{colortbl}
\usepackage{dashrule}
\usepackage{booktabs}
\usepackage{subfigure} 
\usepackage{graphicx}
\usepackage{harpoon}
\usepackage{ragged2e}   
\usepackage{makecell}

\usepackage{threeparttable}  
\usepackage{multirow,booktabs,color,soul,threeparttable}
\definecolor{hl}{rgb}{0.75,0.75,0.75}
\sethlcolor{hl}
\usepackage{float}
\usepackage{etoolbox}

\icmltitlerunning{Context-aware  Diversity Enhancement for Neural Multi-Objective
Combinatorial Optimization}

\begin{document}
\twocolumn[
\icmltitle{Context-aware  Diversity Enhancement for Neural Multi-Objective
Combinatorial Optimization }
%% PSL
\icmlsetsymbol{equal}{*}

\begin{icmlauthorlist}
\icmlauthor{Yongfan Lu}{yyy}
\icmlauthor{Zixiang Di}{yyy}
\icmlauthor{Bingdong Li}{yyy}
\icmlauthor{Shengcai Liu}{comp}
\icmlauthor{Hong Qian}{yyy}
\icmlauthor{Peng Yang}{sch}
\icmlauthor{Ke Tang}{sch}
\icmlauthor{Aimin Zhou}{yyy}
\end{icmlauthorlist}

\icmlaffiliation{yyy}{East China Normal University}
\icmlaffiliation{comp}{A*STAR}
\icmlaffiliation{sch}{Southern University of Science and Technology}

\icmlcorrespondingauthor{Bingdong Li}{bdli@cs.ecnu.edu.cn}

\icmlkeywords{Machine Learning, ICML}

\vskip 0.3in
]

\printAffiliationsAndNotice{\icmlEqualContribution}

\begin{abstract}
Multi-objective combinatorial optimization (MOCO) problems are prevalent in various real-world applications. Most existing neural MOCO methods  rely  on problem decomposition to transform an MOCO problem into a series of singe-objective combinatorial optimization (SOCO) problems and \textcolor{black}{train attention models based on a \textcolor{black}{single-step and} deterministic
greedy rollout.}
% and utilize precise hypervolume to enhance diversity. 
However, \textcolor{black}{inappropriate decomposition and  undesirable short-sighted behaviors of previous methods 
 tend to induce a decline in diversity.} 
To address the above limitation, \textcolor{black}{we design a Context-aware Diversity Enhancement algorithm named CDE, which casts the neural MOCO problems as conditional sequence modeling via autoregression  (node-level context awareness) and
\textcolor{black}{establishes a direct relationship between the mapping of preferences  and diversity indicator of reward} based on hypervolume expectation
maximization (solution-level context awareness).}  
\textcolor{black}{Based on the solution-level context awareness}, we further propose a hypervolume residual update strategy to enable the Pareto attention model to capture both local and non-local information of the Pareto set/front. 
% We also design a novel inference approach to further improve quality of the solution set and speed up hypervolume calculation. 
\textcolor{black}{The proposed CDE can effectively and efficiently grasp the context information, resulting in  diversity enhancement.}
Experimental results on three classic MOCO problems demonstrate that our CDE outperforms several state-of-the-art  baselines.
\end{abstract}

\section{Introduction}
\label{intro}
Multi-objective combinatorial optimization (MOCO) problems \cite{ehrgott2000survey} 
are commonly seen in various fields, such as communication routing \cite{fei2016survey}, 
% investment planning, vehicle routing, 
logistics scheduling \cite{zajac2021objectives}, etc.
Typically, an MOCO problem requires the simultaneous optimization of multiple conflicting objectives,
where the amelioration of an objective may lead to the deterioration of others. It is therefore desirable to discover a set of optimal solutions for MOCO problems, known as 
the Pareto optimal set \cite{yu1974cone}. 
Unfortunately, finding all the Pareto-optimal solutions of an MOCO problem is a challenging task, particularly considering that a single-objective combinatorial optimization (SOCO) problem might already be NP-hard.

% Solving such kind of problems requires taking into account different roles’ preferences corresponding to different objectives, which may often conflict with each other. In principle, the goal of MOCOPs is to find the best compromise solutions (known as Pareto optimal solutions) rather than a single optimal solution. The decision maker can eventually choose a particular Pareto optimal solution according to his knowledge for practical usage.

Due to the exponentially increasing computational time required for exactly tackling MOCO problems \cite{ehrgott2016exact}, heuristic methods have been favored to yield an approximate Pareto set in practice over the past few decades. Although heuristic methods \cite{herzel2021approximation} are relatively efficient, they rely on domain-specific knowledge and involve massive iterative search. Recently, inspired by the success of deep reinforcement learning (DRL) in learning neural heuristics for solving SOCO problems, researchers have also explored DRL-based neural heuristics \cite{li2020deep,lin2022pareto} for MOCO problems. Typically, parameterized as a deep model, these neural heuristics adopt an end-to-end paradigm to construct solutions without iterative search, significantly reducing computational time compared with traditional heuristic methods.

Existing neural MOCO methods typically decompose an MOCO problem into a series of SOCO subproblems by aggregation functions and \textcolor{black}{then train attention models based on a \textcolor{black}{single-step and} deterministic greedy rollout to obtain a Pareto set}. This approach can be seen as an extension of decomposition-based multi-objective optimization strategies \cite{zhang2007moea} to an infinite preference scenario \textcolor{black}{with the aid of REINFORCE}. However, previous methods suffer from several drawbacks. 
\textcolor{black}{Firstly, conventional Markov decision-based neural MOCO algorithms  tend to induce undesirable short-sighted behaviors  because of the neglect of sequence property \cite{rybkin2021model}, \textcolor{black}{which might lead to over-optimization of an objective and fall into local optimum \cite{kamalaruban2020robust}.}
Secondly, \textcolor{black}{the solutions mapped by different subproblems are tackled independently, which harms diversity since only local term is considered in the reward function
and mutual supportiveness of subproblems is largely ignored. 
% and there is a lack of clarity in interpreting the integration of aggregation functions for subproblems.
} 
\textcolor{black}{Both drawbacks result in the loss of diversity.}
% Thirdly, while there exists an algorithm \cite{chen2023neural} that introduced  hypervolume into the reward  to enhance diversity, the precise calculation of hypervolume is time-consuming and  unaffordable due to its NP-hard nature \cite{friedrich2009multiplicative}. 
}
% Thus, it is necessary to context information to enhance diversity effectively and efficiently.

% However,  most existing algorithms tend to separate different preferences without considering the interconnections between subproblems, which causes , and further falls into the predicament of poor diversity. Although there are few algorithms  to introduce  hypervolume into the reward for DRL of MOCO recently, the precise calculation of HV incurs a significant time cost that is deemed unaffordable.

After identifying the above-mentioned challenges, we propose to utilize  context information to enhance diversity effectively and efficiently.
Specifically, we have extended the idea of sequence modeling \cite{chen2021decision,janner2021offline} and hypervolume expectation maximization \cite{deng2019approximating,zhang2020random,zhang2023hypervolume} to MOCO to obtain node- and solution-level context awareness.
% we propose a novel geometry-aware Pareto set learning algorithm for MOCO. 
% Firstly, our method adopts a geometric perspective for problem decomposition, formulating it as a hypervolume expectation maximization problem. 
% This approach establishes a clear correspondence between specific preferences and the resulting solutions of subproblems in a polar coordinate system. 
% Secondly, we propose a hypervolume residual update (HRU) strategy to introduce local and non-local information while training the Pareto attention model.
% % mitigate  the issue of global information loss caused by the varying attention given to different preferences by the Pareto attention model. 
% Lastly, we propose an explicit and implicit dual inference ($\rm{EI^2}$) approach and a local subset selection acceleration strategy, which offer several advantages, including superior convergence and diversity, as well as efficient selection. 
The contributions of this work can be summarized as follows: (1) \textcolor{black}{We design a \textbf{\underline{C}}ontext-aware \textbf{\underline{D}}iversity  \textbf{\underline{E}}nhancement algorithm, termed as CDE, which is the first to put forward the ideas of node and solution-level context awareness for MOCO. Node-level context awareness casts the neural MOCO problems as conditional sequence modeling and autoregressively update node embeddings. Solution-level context awareness establishes a direct relationship between the mapping of preferences and diversity indicator of reward based on hypervolume expectation
maximization.} 
% The solutions derived from CDE are aligned precisely
% with the preferences in a polar coordinate system under mild conditions. 
% Mutual supportiveness of subproblems is the first to be leveraged to conduct geometry-adaptive learning for neural MOCO.
(2) \textcolor{black}{Based on the solution-level context awareness}, we propose a hypervolume residual update (HRU) strategy to make the Pareto attention model grasp both local and non-local information of the PS/PF and avoid being misled by weak solutions to a certain degree.
% (3) We design a novel inference approach named explicit and implicit dual inference ($\rm{EI^2}$) to improve convergence and diversity. 
% (3) We propose an explicit and implicit dual inference ($\rm{EI^2}$) approach, which can improve the convergence and diversity
% of solutions. Moreover, approximate hypervolume calculation and local subset selection acceleration (LSSA) strategy are introduced to facilitate efficient diversity enhancement and selection, respectively.
(3) Experimental results on three classic MOCO problems demonstrate that our CDE outperforms
state-of-the-art neural baselines by superior decomposition and efficient diversity enhancement.

\section{Related Work}

\paragraph{Neural Heuristics for MOCO.} 
Decomposition is a mainstream scheme in learning-based methods for multi-objective optimization \cite{lin2022pareto, navon2020learning, lin2019pareto}. 
% An MOCOP can be decomposed into a series of single-objective CO problems and then solved by neural construction methods to approximate the Pareto set. 
% \textcolor{black}{the two sentence are Repetitive.
% I think the second one is better.}
Their basic idea
is to decompose MOCO problems into multiple subproblems according to prescribed weight vectors, and then train a single model or multiple models to solve these subproblems. 
% \textcolor{black}{Better to use people (Zhang et al.) as 
% the subject of verbs like 'propose' or 'train', instead of xx paper }
For example,
\cite{li2020deep} and  \cite{zhang2021modrl} train 
multiple models collaboratively through a transfer learning strategy. 
% Preference-conditioned multi-objective combinatorial optimization (PMOCO) 
\cite{lin2022pareto} train a hypernetwork-based model to generate the decoder parameters conditioned on the preference for MOCO (PMOCO).
Both MDRL \cite{zhang2022meta} and EMNH \cite{chen2023efficient} leverages meta-learning to
train a deep reinforcement learning model that could be fine-tuned for various subproblems. NHDE \cite{chen2023neural} proposes indicator-enhanced DRL with an HGA model, which is the first to introduce the hypervolume into the reward for MOCO to enhance diversity.
% Decomposition is a mainstream scheme in learning-based methods for multi-objective optimization \cite{lin2022pareto}. 
% Their basic idea
% is to decompose MOCO problems into multiple subproblems according to prescribed weight vectors, and then train a single model or multiple models to solve these subproblems. Details can be found in Appendix \ref{related}.
% For example,
% \cite{li2020deep} and  \cite{zhang2021modrl} train 
% multiple models collaboratively through a transfer learning strategy. 
% \cite{lin2022pareto} train a hypernetwork-based model to generate the decoder parameters conditioned on the preference for MOCO (PMOCO).
% Both MDRL \cite{zhang2022meta} and EMNH \cite{chen2023efficient} leverages meta-learning to
% train a deep reinforcement learning model that could be fine-tuned for various subproblems. NHDE \cite{chen2023neural} proposes indicator-enhanced DRL with an HGA model, which is the first to introduce the hypervolume into the reward for MOCO to enhance diversity.

\paragraph{Hypervolume Expectation Maximization.} 
The hypervolume  indicator measures the quality of MOCO solution sets and is consistent with Pareto
dominance. Maximizing hypervolume is a basic principle in multiobjective optimization algorithm
design \cite{emmerich2005emo, zitzler2007hypervolume}. Many methods leverage hypervolume maximization to train neural networks. Deist et al. \cite{deist2023multi}  adopt the idea of gradient search to obtain a finite set of Pareto solutions. Zhang et al. \cite{zhang2023hypervolume} extends hypervolume scalarization of a finite set \cite{shang2018new} to hypervolume expectation with a simple MLP. However, in terms of sequential prediction problems, traditional MLP is not suitable and the introduction of  hypervolume expectation maximization needs to be further studied.

\textcolor{black}{\paragraph{Sequence Modeling.} Sequence modeling excels in capturing temporal dependencies and structured data\cite{bellemare2013arcade}, providing robust predictions for sequential tasks compared to the trial-and-error nature of traditional Markov decision. 
Chen et al. \cite{chen2021decision} propose a offline method, termed decision transformer, which is the first to cast the problem of RL as conditional sequence modeling and  conditions an autoregressive model on the desired return (reward), past states, and actions. Janner et al. \cite{janner2021offline} proposes trajectory transformer, which further models distributions over trajectories and repurposing beam search as a planning algorithm and demonstrate the flexibility of this approach across long-horizon dynamics prediction, imitation learning and so on.}

\section{Preliminaries}
\subsection{Multi-Objective Combinatorial Optimization}
Without loss of generality,
an MOCO problem
can be mathematically stated as follows :
    \begin{eqnarray}
            \begin{aligned} \label{mopdef}
    &\min \textbf{ } \boldsymbol{ f}(\boldsymbol{x})=(f_{1}(\boldsymbol{x}), f_{2}(\boldsymbol{x}),\ldots, f_m(\boldsymbol{x}))^{T}\\
    &\textrm{ }s.t.\textrm{ }  \boldsymbol{x} \in \mathcal{X}   
    \end{aligned},
  \end{eqnarray} 
where 
$\boldsymbol{x}$ $=$ $(x_1,x_2,\ldots,x_d)$ is the decision vector, 
$\boldsymbol{f} (\cdot) $: $ \mathcal{X} \rightarrow \mathcal{T}$ is 
$m$
objective functions,
$ \mathcal{X} $ and $\mathcal{T}$ denote the (nonempty)  \textit{decision space} 
and  the \textit{objective space}, respectively.  
Since the objectives are usually in conflict
with each other, a set of trade-off solutions is
to be sought. The concept of \textit{Pareto optimality} is introduced.
% MOPs with a large number of decision variables (usually 100) are referred to as 
% large-scale multi-objective optimization problems (LSMOPs) \cite{tian2021evolutionary}.

%pang2022counterintuitive
% }
%
%
 % \textcolor{black}{
% \newtheorem{theorem}{\bf Theorem}
% \newtheorem{definition}{\bf Definition} 
\begin{definition}[Pareto dominance~\cite{yu1974cone}]
	Given two solutions $\boldsymbol{a}$, $\boldsymbol{b}$ in
 the  region 
 % \textcolor{black}{not consistent with above definition. feasible or not?\\
 % better to say 'of a minimization MOP'}
 $ \mathcal{X}$,
$\boldsymbol{a}$ is said to \textit{dominate} $\boldsymbol{b}$ (denoted as $\boldsymbol{a} \prec  \boldsymbol{b}$)
if and only if 
 $\forall i \in \{1,2,...,m\}, f_{i}(\boldsymbol{a}) \leq f_{i}(\boldsymbol{b})$
and $\exists j \in \{1,2,...,m\}, f_{j}(\boldsymbol{a}) < f_{j}(\boldsymbol{b})$.
\end{definition} 
\begin{definition}[Pareto optimality]
    A solution $\boldsymbol{a}^{\ast}  \in \mathcal{X}$
is  Pareto optimal
if no other solution $\boldsymbol{a}  \in \mathcal{X}$
  can dominate it.
\end{definition} 
\begin{definition}[Pareto Set and Pareto Front]
    The  solution set consisting of all the Pareto optimal solutions is called the \textit{Pareto set} (PS): 
$PS$$=$$\{\boldsymbol{a}  \in \mathcal{X}| \forall \boldsymbol{b}  \in \mathcal{X},\boldsymbol{b} \not\prec  \boldsymbol{a}   \}$
and 
  the corresponding objective vector set of the PS is the \textit{Pareto front} (PF). 
\end{definition} 

% \subsection{Aggregation Function}
% An aggregated  (or utility) function can map each point in the objective space
% into a scalar according to an $m$-dimensional weight vector $\lambda$ with $\|\boldsymbol{\lambda}\|_p=1$ ( $l_p $-norm constraint ) and $\lambda_i \geq 0$. Weighted-Sum (WS) and Weighted-Tchebycheff are commonly used utility functions \cite{miettinen1999nonlinear}. As the simplest representative, WS can be defined by $\min_{\boldsymbol{x} \in \mathcal{X}} f(\boldsymbol{x}|\boldsymbol{\lambda})=\sum^M_{i=1} \lambda_i f_i(\boldsymbol{x})$.
% \textcolor{black}{Is is necessary to introduce agg func
% as an subsection? }
\subsection{Performance Evaluation}
The nadir/ideal point of an MOCO problem is constructed by the worst/best objective values of the Pareto
set
$y^{nadir}_i={sup}_{\boldsymbol{y} \in \mathcal{T}}\{y_i\}$  and $y^{ideal}_i={inf}_{\boldsymbol{y} \in \mathcal{T}}\{y_i\}$, $\forall i \in 1, ..., m$.  
Given a set of objective
vectors $Y$,
the hypervolume (HV) indicator \cite{zitzler1999multiobjective} is defined 
% as a metric of the optimality of  detailed 
as follows,
     \begin{eqnarray}
                \mathcal{HV}_r(Y) = \Lambda(\{\boldsymbol{q} \in \mathbb{R}^d| \exists \boldsymbol{p} \in Y:\boldsymbol{p} \preceq \boldsymbol{q}\; \textbf{and}\; \boldsymbol{q} \preceq \boldsymbol{r} \}),
                \label{hvv}
            \end{eqnarray}
where $\Lambda(\cdot)$ denotes the Lebesgue measure. and $\boldsymbol{r}$ is a reference vector. 
% \textcolor{black}{citation of HV original paper?}
We require that
$\boldsymbol{r}$ $\succeq$ $\boldsymbol{y}^{nadir}$. All methods share the same reference point $\textbf{r}$ for a problem (see Appendix \ref{refer}).

\section{Methodology}
\subsection{Overview}
\begin{figure*}[!htbp]
			\centering

                \includegraphics[width=0.95\textwidth]{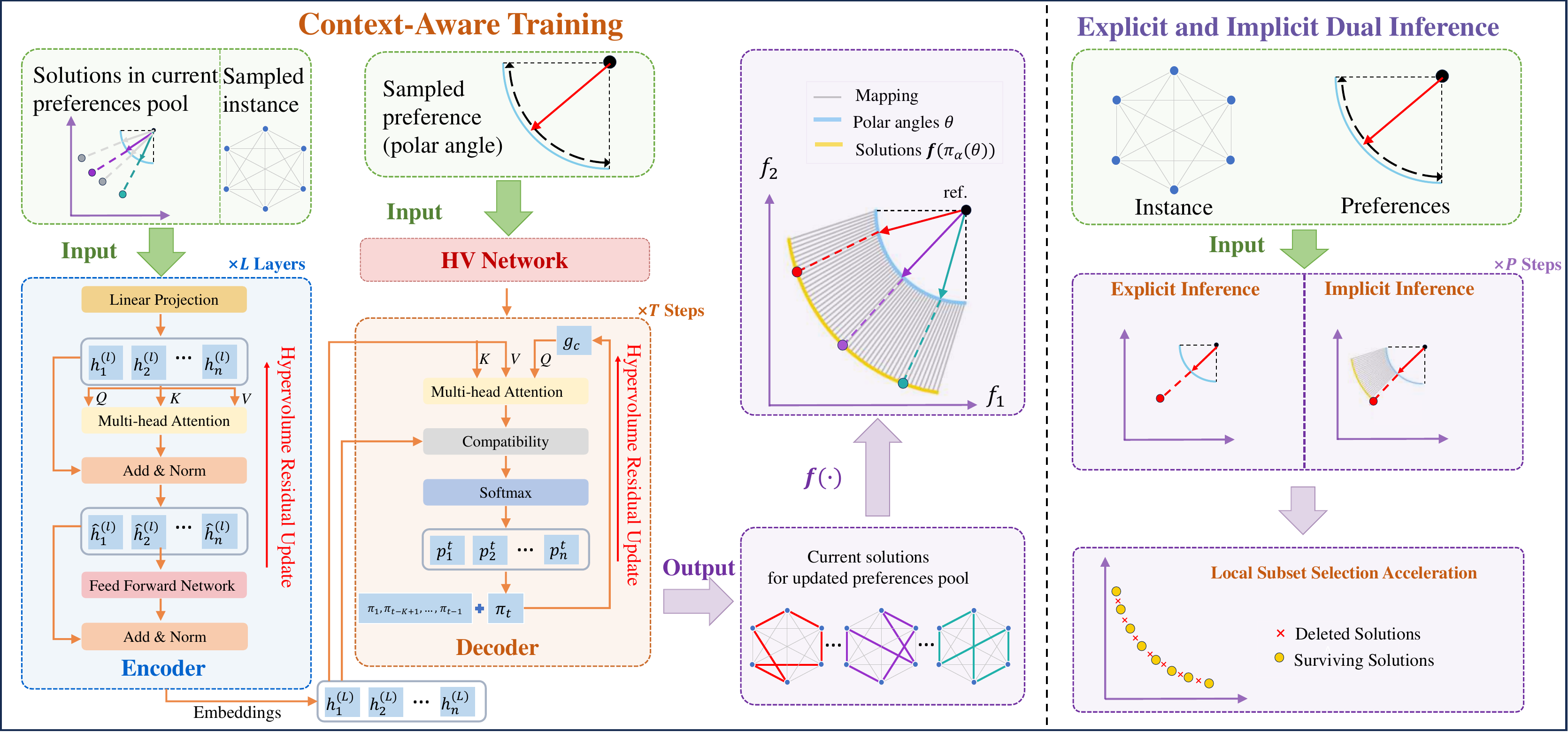}
                \caption{Pipeline of  CDE. \textbf{Left} ( $p'$-th training iteration): \textcolor{black}{CDE is trained using a context-aware strategy to establish the mapping between preferences (polar angles) and solutions with effective and efficient diversity enhancement. Specifically, the encoder takes in a sampled instance for projection and solutions generated from previous preferences $\boldsymbol{\theta}^{p‘-1}$ offer context information for the current preference $\theta$. In addition, $\theta$ generates the decoder parameters through a specifically designed HV network. At step $t$ in the decoder, a context embedding $\boldsymbol{g}_c$ is used to calculate the probability of node selection.} The solutions derived from CDE are aligned precisely with the polar angles in a polar coordinate system under mild conditions. \textbf{Right}: during inference, CDE employs an explicit and implicit dual inference ($\rm{EI^2}$) approach to enhance convergence and diversity, and incorporates local subset selection acceleration (LSSA) to enhance efficiency.  A well-trained Pareto attention model applies the $\rm{EI^2}$ approach to each polar angle, resulting in solutions with superior convergence and diversity, while LSSA facilitates efficient selection.}
                \label{flow}
                \vspace{-0.1cm}
\end{figure*}
\label{submission}
% \textcolor{black}{ first idea and how to do it;
% then give its advantages. Follows the zong-fen structure.\\ Instead of current style:
% our method good (in the beginning). Readers will 
% wonder, why? what happened?
% \\
% The advantages sentences need to be moved to other places.}

\textcolor{black}{Our context-aware diversity enhancement (CDE) takes the first attempt to cast the neural MOCO problems as conditional sequence modeling and and establishes a direct relationship between the mapping between preferences and solutions and hypervolume based on hypervolume expectation
maximization (HEM) for neural MOCO,   enhancing diversity via context-aware from two aspects (Figure \ref{flow} and Algorithm \ref{gadpsl}). }
% We extend the concept of hypervolume scalarization from a finite set \cite{zhang2020random} to the hypervolume expectation maximization of the approximate infinite Pareto set/front.
% Specifically, we extend hypervolume scalarization  of a finite set to hypervolume maximization expectation of the approximate Pareto infinite set. 
% Then we design a geometry adaptive structure which offer a  geometric perspective for the mapping between the preferences and solutions (Figure \ref{flow} and Algorithm \ref{gadpsl}).
% CDE provides .
% In addition, we design a hypervolume residual update strategy to deal with the predicament that weak solutions may make Pareto
% attention model misjudge the global information when the focus of
% Pareto attention model to different preferences is inequable. 
% In order to overcome the predicament of weak solutions potentially leading to misjudgment of global information by the Pareto attention model when there is an unequal focus on different preferences, we design a hypervolume residual update strategy.
Moreover, we  design a hypervolume residual update (HRU) strategy
to better grasp local and non-local information and avoid the potential misleading caused by  
the unequal focus of different preferences.
We also introduce an explicit and implicit dual inference ($\rm{EI^2}$) approach that enhances quality and efficiency, supported by local subset selection acceleration (LSSA).

\subsection{Node-level Context Awareness via Sequence Modeling}
Given a problem instance $s$, we sequentially solve its subproblem $i \in \{1,...,P\}$. The $i$-th subproblem is associated with weight polar angle $\theta^i$, which follows uniform distribution Unif($\Theta$): $\Theta = [0,\frac{\pi}{2}]^{m-1}$. Let $\boldsymbol{\pi}^i=(\pi^i_1,...,\pi^i_T)$ denotes the obtained solution at step $i$. 
Given a fully connected graph
of $T$ nodes (cities) with $m$ cost objectives of an MOTSP instance, $\boldsymbol{\pi}$ denotes a tour that visits each city exactly once and returns to the starting node, and  the objective functions are an
 $m$-dimensional vector $\boldsymbol{y}(\boldsymbol{\pi}) = (y_1(\boldsymbol{\pi}),...,y_m(\boldsymbol{\pi}))$.

The idea of sequence modeling has been applied in many fields. However, there was no attempt to integrate sequence modeling into MOCO previously. In this paper, we conduct the first attempt. We follow the idea of encoder in PMOCO except the update of node embeddings \{$\boldsymbol{h}$$_{\boldsymbol{\pi}_1}$,...,$\boldsymbol{h}$$_{\boldsymbol{\pi}_n}$\} in each layer. The decoder models trajectories autoregressively with minimal modification to the decoder in PMOCO. At time step $t$ in MOTSP, the context trajectory consists of two parts: 1) the first selected node embedding $\boldsymbol{\pi}_1$; 2) the nearest $K$ nodes embeddings \{${\boldsymbol{\pi}_{t-K}}$,...,${\boldsymbol{\pi}_{t-1}}$\}.  Then, the sequence embedding is represented as \{$\boldsymbol{h}$$_{\boldsymbol{\pi}_1}$, $\boldsymbol{h}$$_{\boldsymbol{\pi}_{t-K}}$,...,$\boldsymbol{h}$$_{\boldsymbol{\pi}_{t-1}}$\}. The context information of other problems can be found in Appendix \ref{Embedding}.  Eq. \ref{hvv1} defines the attention score, which is normalized as $\tilde\alpha$ by softmax, 

     \begin{eqnarray}
     \begin{aligned}
                \alpha_{uv} =& \frac{(W^Q (\boldsymbol{h}_{\boldsymbol{\pi}_u}+PE(u)))(W^K(\boldsymbol{h}_{\boldsymbol{\pi}_v}+PE(v))^T}{\sqrt{d/H}}\\
                &+ f(\boldsymbol{h}_{\boldsymbol{\pi}_u}+PE(u), \boldsymbol{h}_{\boldsymbol{\pi}_v}+PE(v)),
                %+ \textnormal{Relu}(W^Q \boldsymbol{h}_{\boldsymbol{\pi}_u}W^A)} 
                \label{hvv1} 
                \end{aligned} 
            \end{eqnarray}
\vspace{-0.5cm}
     \begin{eqnarray}
     \begin{aligned}
\tilde{\boldsymbol{h}}_{\boldsymbol{\pi}_t}=\sum_{v=1}^{n}(\tilde\alpha_{1v}W^V\boldsymbol{h}_{\boldsymbol{\pi}_v}+\sum_{u=t-K}^{t-1} \tilde\alpha_{uv}W^V\boldsymbol{h}_{\boldsymbol{\pi}_v}).
                \label{hvv2}
                                \end{aligned} 
            \end{eqnarray}
where $PE(\cdot)$ denotes the position embedding, which is important to the sequence modeling. Moreover, $f(\cdot)$ denotes the feedforward neural network, which is introduced to generate a dynamic attention weight of current sequence element and key. \textcolor{black}{The above practice is essential to neural MOCO since the importance of previous node position and order for the next node prediction varies wth the change of instance, iteration and so on.} Finally, as for the multi-head attention (MHA), $\tilde{\boldsymbol{h}}_t$ is further computed as follows,
\begin{eqnarray}
                \tilde{\boldsymbol{h}}_{\boldsymbol{\pi}_t}=W^OConcat(\tilde{\boldsymbol{h}}_{\boldsymbol{\pi}_{t,1}},...\tilde{\boldsymbol{h}}_{\boldsymbol{\pi}_{t,H}}),
                \label{hvv3}
            \end{eqnarray}
where $\tilde{\boldsymbol{h}}_{\boldsymbol{\pi}_t}$ for head $h \in \{1,...H\}$ is  obtained according to Eq. \ref{hvv2}, which is used for update the node embeddings of the last layer by $ \textrm{Add} \& \textrm{Norm}$ and $\textrm{FFN}$. In the traditional MHA, $W^Q$, $W^K$, $W^V$ and  $W^O$ are independent trainable parameters. Besides, the dynamic learnable parameters $W^A$ is introduced to improve the sensitivity to context information (i.e. attention to the $K$ different nodes) when predicting the current node. The practice can effectively improve 1) contextual relevance; 2) flexibility and adaptability; 3) learning efficiency; 4) robustness to variability \cite{chen2021decision}.
\subsection{Solution-level Context Awareness via Hypervolume Expec-
tation Maximization}
\textcolor{black}{Pareto set learning (PSL) is the mainstream idea of dealing with MOCO currently \cite{lin2022pareto}, which   originates from traditional optimization \cite{hillermeier2001nonlinear} and aims to
learn the whole Pareto set by a model. }
% The core idea of PSL is based on  decomposition, which is consistent with basic idea of neural MOCO.
Current neural MOCO methods tend to leverage weighted-sum (WS) or weighted-Tchebycheff (TCH)  aggregation and/or use accurate HV to balance convergence (optimality) and diversity \cite{lin2022pareto, chen2023neural}. 
However, they may overemphasize convergence or improve diversity at the sacrifice of
high computational overhead.
 % time-consuming. 
\textcolor{black}{Thus, we propose a novel solution-level context awareness to formulate PSL based on hypervolume expectation maximization, which is only used with MLP in \cite{zhang2023hypervolume} before.}

Here  an HV scalarization function is introduced for unbiased estimation of 
% a connection between
% the scalarization of  a finite set and 
the hypervolume indicator:
\begin{lemma}[Hypervolume scalarization of a finite set \cite{shang2018new,deng2019approximating,zhang2020random}] 
\label{lemma1}
$Y=\{\boldsymbol{y}^1,...,\boldsymbol{y}^n\}$ denotes set of finite objective vectors and $\boldsymbol{r} = (r_1,...r_m)$ denotes a
reference point, which satisfies $\forall i \in \{1,2,...,m\}, \boldsymbol{r} \succ \boldsymbol{y}^i$.
\begin{eqnarray} 
\label{hv}
            \begin{aligned}
   &\mathcal{HV}_{\boldsymbol{r}}(Y) := \frac{\Phi}{m 2^m} \mathbb{E}_{\theta}[(\max_{\boldsymbol{y} \in Y}\{\mathcal{G}^{mtch}(\boldsymbol{y}| \theta, \boldsymbol{r}) \})^m]\\
   &\mathcal{G}^{mtch}(\boldsymbol{y}| \theta, \boldsymbol{r})=\min_{i \in \{1,...,m\}}\{(r_i-y_i)/\lambda_i(\theta)\}
              \end{aligned},           
            \end{eqnarray}
where $\Phi=\frac{2\pi^{m/2}}{\Gamma(m/2)}$ denotes the area of the $(m-1)$-D unit sphere, which is a dimension-specific constant and $\Gamma(x)=\int^{\inf}_0 z^{x-1}e^{-z}dz$ denotes the analytic continuation of the
factorial function. $\boldsymbol{\lambda}(\theta)=(\lambda_1(\theta),...,\lambda_m(\theta))$ is the preference vector. It is the Cartesian coordinate of $\theta$ on the positive unit sphere $\mathbb{S}_{+}^{m-1}$:
% , which can be obtained as follows
% \begin{align}\left\{\begin{aligned}
$\lambda_1(\theta) = \sin\theta_1\sin\theta_2\ldots\sin\theta_{m-1},
        \lambda_2(\theta) = \sin\theta_1\sin\theta_2\ldots\cos\theta_{m-1},
        ...,
        \lambda_m(\theta) = \cos\theta_1$.
% \end{aligned}.\right.\end{align}
\end{lemma}

This scalarization function is a computationally-efficient
and accurate estimator of the hypervolume  that
easily generalizes to higher dimensions. 
Moreover, it is polynomial in the number of objectives $m$ when $\mathcal{X}$ is suitably compact. This implies that the choice of scalarization functions and
 polar angles distribution Unif($\Theta$) are theoretically sound and leads to
provable convergence to the PF.

% Note the resulting hypervolume scalarization functions come with provable guarantees. 
Given a set of uniform polar angles, accurately estimating HV with diminutive variance relies on numerous uniformly distributed solutions.
% However, expensive Monte Carlo estimation  of the  finite set $Y$  for the HV 
However, solutions are not evenly distributed, especially in the nascent stage of optimization.
% exploration and exploitation.
 Thus, it is necessary to build a mapping between the preferences (polar angles) and the PS/PF, replacing the finite set $Y$ via a Pareto attention model. In the other words, the Pareto attention model is able to predict nodes (cities) sequentially when given embedded preferences and further maps the node sequence from the instance space to the objective space. To sum up, what we need to realize is the mapping from preference space to objective space, with instance space as a ``bridge''. Thus, we follow the idea of hypervolume expectation maximization \cite{zhang2023hypervolume} to construct our Pareto attention model. The detailed method of extending  the scope of the input set in Lemma \ref{lemma1} to the  PF containing an infinite number
of objective vectors refers to Appendix \ref{extension}.

Based on the above discussion, it is favorable and feasible to take into account estimated hypervolume \textcolor{black}{with context awareness} when training our Pareto attention model $\Psi_{\beta}(\theta, s)$. The update of the model will be discussed thoroughly in the next subsection.
% by maximizing a composite reward (Eq. \ref{reward}), which is constructed by projection distance  and estimated hypervolume, respectively. 
% The reward will be discussed in the next subsection. 
% Here, we just discuss alternative unbiased calculation of the HV. 
% In terms of non-local term $\mathcal{HV}_{\boldsymbol{r}}(Y)$, 
One remaining issue is how to generate the preference-conditioned parameters $\beta_{decoder}(\theta)$. We design an HV Network with reference to hypernetwork \cite{schmidhuber1992learning, shamsian2021personalized, sarafian2021recomposing}, which provides a powerful and efficient way for conditional computation and is widely used
in transfer learning \cite{ying2018transfer}. We use a simple MLP: $\beta_{decoder}(\theta)$ = MLP$(\lambda|\psi)$  to generate the decoder parameters conditioned on the polar angles, which is the first to take estimated HV into account. In view of this, our method can better capture the global characteristics of the PF. Moreover, our approximate calculation of hypervolume can better reduce time consumption and enhance model robustness. More details can be found in 
Appendix \ref{Model}. 
We also derive an alternative form of CDE which uses the ideal point as a reference point and obtains interesting results in Appendix \ref{gaplalter}.

\begin{algorithm} [!htbp]
    % \LinesNumbeblack
	\caption{The training process of CDE} \label{gadpsl} 
\begin{algorithmic}[1]
    % \SetKwData{Q}{Q}
    % \SetKwInOut{Input}{input}\SetKwInOut{Output}{output}
   \STATE {\bfseries Input:} preference distribution $\Theta$,instance distribution $\mathcal{S}$, number of training steps $E$, maximal size of polar angles pool $P'$,  batch size $B$, instance size $N$, length of sequence $K$
     \STATE Initialize the model parameters $\beta$   
    \FOR{$e=1$ {\bfseries to} $E$}
    \STATE  $s_i$ $\sim$ {\bfseries SampleInstance} ($\mathcal{S}$) \quad $\forall i$ $ \in \{1,...B\}$ 
     \STATE  Initialize $\mathcal{F}^j_i$ $\leftarrow$ $\emptyset$ \quad $\forall i
    \forall j \in \{1,...N\}$  
     \STATE  Initialize $\boldsymbol{\theta}^0$ $\leftarrow$ $\emptyset$ 
    \FOR{$p'=1$ {\bfseries to} $P'$}
        \STATE   $\theta$ $\sim$  {\bfseries  SamplePreference}($\Theta$) 
         \STATE  $\boldsymbol{\theta}^{p'}$ $\leftarrow$ $\boldsymbol{\theta}^{p'-1} \cup \theta$ 
       \STATE    $\pi^j_i$ $\sim$ {\bfseries  SampleSolution}($Prob_{\beta(\theta)}(\cdot|s_i,\theta),\mathcal{F}^j_i$, $K$) 
       \STATE    $\mathcal{V}_{\beta}(\theta,s)$,$\widetilde{\mathcal{HV}_r}(\beta, \boldsymbol{\theta}^{p'}, s)$ $\leftarrow$ Calculate the projection distance and approximate HV by Eq. \ref{comput} $\forall i,j$ 
        \STATE   $\alpha^j_i$ $\leftarrow$ Calculate expected HV improvement by Eq. \ref{ehvi} \quad $\forall i,j$ 
        \STATE  $R^j_i$ $\leftarrow$ Calculate the reward by Eq. \ref{reward} \quad $\forall i,j$ 
        \STATE   $b_i$ $\leftarrow$ $\frac{1}{N} \sum^n_{j=1}$($R^j_i$) \quad $\forall i$ 
         \STATE  $\nabla \mathcal(J)$($\beta$) $\leftarrow$ $\frac{1}{Bn} \sum^B_{i=1} \sum^n_{j=1}[(-R^j_i$ 
          \STATE       $\qquad \qquad$ $-b_i)\nabla_{\beta(\theta)}\log Prob_{\beta(\theta)}(\pi^j_i|s_i,\theta,\mathcal{F}^j_i)]$
         $\beta$ $\leftarrow$ {\bfseries  Adam}($\beta,\nabla \mathcal(J)(\beta)$) 
      \STATE   $\mathcal{F}^j_i$ $\leftarrow$ $\mathcal{F}^j_i$ $\cup$ $\{\boldsymbol{f}(\pi^j_i)\}$ \quad $\forall i,j$ 
     \ENDFOR
    \ENDFOR
    \STATE {\bfseries Output:} The model parameter $\beta$
    \end{algorithmic}
\end{algorithm}

\subsection{Hypervolume Residual Update Strategy}
% In order to make our novel Pareto attention model show superior performance, 
\textcolor{black}{Solution-level context awareness} via hypervolume expectation
maximization is based on the assumption of unequal importance for all subproblems. The assumption further requires the model to pay different attention to each subproblem. 
In order to enable the Pareto attention
model to capture both local and non-local
information of the Pareto set/front,
we propose a hypervolume residual update (HRU) strategy to conduct more effective training.
% In terms of 
The reward function of subproblem $i$ is defined as follows:
\begin{eqnarray} 
\label{reward}
            \begin{aligned}
   R_i= \omega \cdot \underbrace{\mathcal{V}_{\beta}(\theta^i,s)}_{local} + \alpha \cdot  \underbrace{\widetilde{\mathcal{HV}_r}(\beta,\boldsymbol{\theta}^i,s)}_{non-local}
              \end{aligned}.           
            \end{eqnarray}
% Specially,
Specifically,
the reward function is composed of
local and non-local terms:
the projection distance of the $i$-th solution with respect to the polar angle $\theta^i$ 
and the
approximate hypervolume of all the selected solutions of subproblems 1 to $i$. 
The advantage of this combination is that it can balance exploration and exploitation. Each preference update will drive the evolution of the subsequent preferences. However, there is a deviation between the mapping solution of the Pareto attention model  and the true solution with given preference and the deviation will decrease with the model training. Therefore, we introduce a local term decay parameter $\omega$, which is set to $1-\frac{ep}{EP}$, where $ep$ denotes the current epoch and $EP$ denotes the total epochs since the local term is easier to learn than the non-local term. With the iteration of training epochs, our reward function will approach HV.
During the training process,  weak solutions may make the Pareto attention model misjudge the global information of the Pareto set/front.
Based on the above considerations, we propose a novel definition, namely estimated hypervolume improvement (EHVI), which is as follows:
\begin{align} 
\label{ehvi}
            \begin{aligned}
   \textbf{EHVI}(\beta,\theta^i,\boldsymbol{\theta}^i,s)= &\widetilde{\mathcal{HV}_r}(\beta,\boldsymbol{\theta}^i,s) 
-\widetilde{\mathcal{HV}_r}(\beta,\boldsymbol{\theta}^i\backslash\{\theta^i\},s).
              \end{aligned}           
            \end{align}
When \textbf{EHVI}($\cdot$) is negative, we infer that the 
current polar angle $\theta^i$ is mapped to a weak solution. In this case, the model will disregard the non-local term in order to 
% prioritize the convergence of the current weak solution and 
avoid the weak solution from misleading the model.  On the other hand, when EHVI($\cdot$) is positive, the non-local term is utilized to transfer valid global information to subsequent solutions. Regarding the above, $\alpha=\lceil \textbf{EHVI} \rceil$ since $\widetilde{\mathcal{HV}_r}$ is a normalized value and then $\textbf{EHVI} \in (-1,1)$.

\subsection{Novel Inference Strategy}
\label{novel}
\paragraph{Explicit and Implicit Dual Inference ($\rm{EI^2}$).}
In the inference phase, for $P$ given preferences, the well-trained model is used
to sequentially solve $P$ corresponding subproblems, as shown in Figure \ref{flow}. 
In previous preference-conditioned inference approaches, each subproblem is treated separately, and the WS/TCH method is employed to determine the optimal solution.
We refer to this approach as explicit inference, as it only considers the current preference directly, 
% disregards the information from surrounding subproblems and previous preferences 
often resulting in local optima and producing duplicate solutions for similar preferences.
To handle these issues, we design an additional implicit inference module,  which aims to maximize \textbf{EHVI} (Eq. \ref{ehvi}). Specifically, we infer all the previous subproblems can provide valuable prior information and construct a partial approximate PF for HV estimation.
The \textbf{EHVI} serves as a suitable indicator to estimate the relevance and importance of the solution of the current preference based on all the previous preferences, effectively reducing duplicate solutions as \textbf{EHVI} tends to zero in such a case.
However, this approach may also carry the risk of local optima, as the distribution of subsequent preferences and solutions is unknown.
To mitigate this,  we propose the explicit and implicit dual inference ($\rm{EI^2}$) approach to enable the cooperation of both inference strategies, thereby facilitating better approximation of the PS/PF.

% In order to make the process of inference more reasonable, we design an explicit and implicit dual inference ($\rm{EI^2}$) approach. The explicit inference is similar to previous methods, which just consider current preference. However, this method are likely to produce same solution with similar preferences when model is not well-trained since each solution is locally optimal. In view of it, we introduce implicit inference, which aims to maximize expected hypervolume improvement, which take all the previous preferences into consideration with a geometry aspect. Two  inference strategies cooperate with each other, hence, better approximate Pareto set/front can be constructed. 
% Finally, local subset selection acceleration is involved to keep the size of solution set, and the details can be found in the next subsection.

\paragraph{Local Subset Selection Acceleration (LSSA).}
\label{44}
After the $\rm{EI^2}$ approach, several solutions with superior convergence and diversity are obtained. Then, we want to obtain the optimal subset with a given size restriction (usually $P$). We introduce 
% potential energy concept from physics 
efficient LSSA \cite{wang2022enhancing} instead of traditional LSS.
LSSA is based on potential energy from the physics domain.
The potential energy of a system $Q$ is defined as follows:
   $E(Q)=\sum_{\boldsymbol{x}^i \in Q} \sum_{\boldsymbol{x}^j \in P \backslash{\boldsymbol{x}^i}} U(\boldsymbol{x}^i,\boldsymbol{x}^j), U(\boldsymbol{x}^i,\boldsymbol{x}^j)=\frac{1}{\|{\boldsymbol{x}^i-\boldsymbol{x}^j}\|^c}$,
where $\|\cdot\|$ denotes the module and $c$ is a control parameter.
Suppose there are $k$ iterations on average for LSS, the overall time complexity
in terms of potential energy calculation is $O(n_1^4mk)$, where $n_1$ and $m$ denote the size of $Q$ and objective dimensions of each solution.

Since the overall time complexity is too high,
an acceleration strategy is adopted by using the properties of
both optimization and the potential energy function.
Let $Q_1$ contain current selected solutions while $Q_2$ contains
the unselected ones. The contribution of a selected one
$e1(\boldsymbol{x})$ and an unselected one $e2(\boldsymbol{x})$ to  $Q_1$ can be defined by $e1(\boldsymbol{x}) = \sum_{\boldsymbol{y} \in Q_1 \backslash \boldsymbol{x}} U(\boldsymbol{x},\boldsymbol{y})$, $e2(\boldsymbol{x}) = \sum_{\boldsymbol{y} \in Q_1} U(\boldsymbol{x},\boldsymbol{y})$. Thus,  $E(Q_1)$ 
% (Eq. \ref{Energya})
is equivalent to $ \sum_{\boldsymbol{x} \in Q_1} e1(\boldsymbol{x})$,
% \begin{align}\left\{\begin{aligned}
%     \label{gg}
%         &e1(\boldsymbol{x})=\sum_{\boldsymbol{y} \in P \backslash \boldsymbol{x}} U(\boldsymbol{x},\boldsymbol{y})\\
%         &e2(\boldsymbol{x})=\sum_{\boldsymbol{y} \in P} U(\boldsymbol{x},\boldsymbol{y})\\
% \end{aligned}\right.\end{align}
% Thus, induced from Eq.\ref{Energya}  and Eq.\ref{gg}, the following equality
% always holds:
% \begin{align}\begin{aligned}
%     \label{gg1}
%         &E(P) = \sum_{\boldsymbol{x} \in P} e1(\boldsymbol{x})
% \end{aligned}\end{align}
Consider one of $Q_1's$ neighbors $Q_1 \backslash \{\boldsymbol{x}\}\cup\{\boldsymbol{y}\}$. 
% where $\boldsymbol{x} \in Q_1$ and $\boldsymbol{y} \in Q_2$. 
% Instead of the time-consuming calculation of Eq. \ref{Energya}. 
we can obtain the difference between the energy values
 of the two systems by efficient LSSA strategy
 % to calculate the energy of a new system 
 % with high efficiency:
% The energy of $Q_1 \backslash \{\boldsymbol{x}\}\cup\{\boldsymbol{y}\}$ is as follows:
% \begin{align}\begin{aligned}
%     \label{gg2}
%         &E(P \backslash \{\boldsymbol{x}\}\cup\{\boldsymbol{y}\}) = E(P)-e1(\boldsymbol{x}) + e2(\boldsymbol{y})-U(\boldsymbol{x},\boldsymbol{y}) 
% \end{aligned}\end{align}
% It is clear that when evaluating the neighbor of $P$ by replacing
% $\boldsymbol{x}$ by $\boldsymbol{y}$, it is not necessary to calculate the energy of the new
% system. Instead, only the difference between the energy values
% of the two systems is needed:
% \begin{align}\begin{aligned}
%     \label{gg3}
        $\Delta E(Q_1)=e1(\boldsymbol{x}) - e2(\boldsymbol{y})+U(\boldsymbol{x},\boldsymbol{y})$.
% \end{aligned}\end{align}
% In the same way, we can also store the contributions difference of all the selected and unselected solutions in the running process. 
The time complexity to update
$E(Q_1)$ is $O(1)$ and
% . Moreover, if all $U(\boldsymbol{x}, \boldsymbol{y})$ are preprocessed, 
the time complexity of LSS
reduces from $O(n_1^4mk)$ to $O(n_1^2m + n_1^2k)$. LSSA strategy can find the optimal subset from the solutions selected by $\rm{EI^2}$ with high efficiency. All details and proofs are provided in Appendix \ref{detailslssa}.

\section{Experimental Study}
\label{exp}
\paragraph{Problems.}
We  conducted comprehensive experiments  to evaluate CDE on 3 typical MOCO problems that are commonly
studied in the neural MOCO field \cite{li2020deep,lin2022pareto}, namely  multi-objective traveling salesman problem (MOTSP) \cite{lust2010multiobjective}, multi-objective capacitated vehicle routing problem (MOCVRP) \cite{lacomme2006genetic},
and multi-objective knapsack problem (MOKP) \cite{bazgan2009solving}. Following the convention in \cite{lin2022pareto}, we consider the instances with different sizes,
i.e. 
$T=20/50/100$ for bi-objective TSP
(Bi-TSP), tri-objective TSP (Tri-TSP) and bi-objective CVRP (Bi-CVRP), and $T=50/100/200$ for bi-objective KP (Bi-KP).

\paragraph{Hyperparameters.} We randomly generated 5000 problem instances on the fly in each epoch for training CDE.
The optimizer is ADAM with learning rate $\eta=10^{-4}$  and weight decay $10^{-6}$
for 200 epochs. \textcolor{black}{The length of sequence $K$ is set to 3/5/8/10 for 20/50/100/200 instance nodes.}
During training, the maximal size of polar angles pool $P'$ is set to 20 for each instance. 
During inference, we generate $P=101$ and $P=10201$ uniformly distributed preferences (polar angles) for $m=2$ and
$m=3$, respectively. Besides, we set the control parameter $c$ to $2m$ and set $N$ to $T$, following \cite{kwon2020pomo}. All hyperparameter studies can be found in Appendix \ref{Hyperparameter}.

\paragraph{Baselines.} We compare  CDE with two kinds of state-of-the-art baselines: 1) Traditional heuristics. We introduce a widely-used evolutionary algorithm  for MOCO: NSGAII \cite{deb2002fast} is a Pareto dominance-based multiobjective genetic algorithm;
% and MOEA/D \cite{zhang2007moea} is a decomposition-based multiobjective evolutionary algorithm. 
LKH \cite{tinos2018efficient} and dynamic programming (DP) are employed to solve the weighted-sum (WS) based subproblems for MOTSP and MOKP, denoted as WS-LKH and
WS-DP, respectively;  PPLS/D-C \cite{shi2022improving} is a specialized MOEA for MOCO with local search techniques.
2)  Neural heuristics. DRL-MOA \cite{li2020deep} decomposes a MOCO with different preferences and builds a Pointer Network \cite{vinyals2015pointer} to solve each subproblem; PMOCO \cite{lin2022pareto} is the first to  train 
a single
model for all different preferences simultaneously. NHDE-P \cite{chen2023neural} proposes an indicator-enhanced deep reinforcement learning method to guide the model, which is the first to take the diversity into consideration in the reward.
EMNH \cite{chen2023neural} leverages 
meta-learning to train a DRL model
that could be fine-tuned for various subproblems. 

% All the experiments
% are conducted with an RTX 4090 GPU and a 1.5GHz AMD EPYC 7742 CPU. 

\paragraph{Metrics.} 
Hypervolume (HV) \cite{audet2021performance} is used for performance evaluation,
% Solution set
% quality is mainly measured by hypervolume (HV) ,
where a higher HV means a better solution set. The average HV, gaps with respect to CDE, and total running time
for 200 random test instances are reported. The best (second-best) results
% and its statistically insignificant
% results at 1\% significance level of a Wilcoxon rank-sum test 
are highlighted in \textbf{bold} (\underline{underline}).
All the compared  methods share the same reference points 
(see Appendix \ref{refer}).

\subsection{Results and Analysis}
The overall comparison results 
% for MOTSP, MOCVRP, and MOKP 
are recorded  in Table \ref{t1},  including the average
HV, gap, and total running time for  200 random test instances. Given the same number of preferences, CDE significantly surpasses PMOCO and NHDE-P across all problems and sizes.
When
instance augmentation (aug.) is equipped, CDE achieves the smallest gap among the methods on most cases except Bi-CVRP100 where DRL-MOA performs better. However, DRL-MOA incurs significantly more training time overhead to prepare multiple models for different weights. 
Besides, DRL-MOA
exhibits significant advantages on problems with extremely imbalanced scales (i.e. Bi-CVRP), as it can more effectively decompose subproblems by utilizing prior information.
This advantage becomes more pronounced {on} instances with larger sizes.
In contrast, preference-conditioned methods rely on uniform acquisition of preferences.  
Anyway, CDE shows superior performance than all the other preference-conditioned methods (e.g. PMOCO and NHDE-P) via its context awareness on Bi-CVRP100. Finally, we believe that CDE has the potential to be further improved by incorporating additional prior information.

%and the performance of CDE  may be further improved if more prior information of imbalanced scales is provided.
Regarding the inference efficiency, CDE  generally requires slightly more time compared with PMOCO due to the utilization of LSSA. However, the LSSA demonstrates a notable characteristic of time complexity that remains unaffected by the number of instance nodes, resulting in high efficiency even for larger instances. Furthermore, the time consumption of CDE is significantly lower than that of NHDE-P, since NHDE-P adopts accurate HV calculation and multiple Pareto optima strategy. This is also observed during the training process.
\begin{table*}[!t]
\caption{Experimental results on 200 random instances for MOCO problems.  }
\centering
\scriptsize
\label{t1}
\begin{tabular}{l|ccc|ccc|ccc}
% \centering
\toprule
                      & \multicolumn{3}{c|}{Bi-TSP20}               & \multicolumn{3}{c|}{Bi-TSP50}               & \multicolumn{3}{c}{Bi-TSP100} \\
Method                   & HV $\uparrow$        & Gap $\downarrow$& Time $\downarrow$ & HV $\uparrow$       & Gap $\downarrow$ & Time $\downarrow$ & HV $\uparrow$      & Gap $\downarrow$  & Time $\downarrow$  \\ \midrule
WS-LKH (101 pref.)       &  0.6268         & 1.10\%   & 4.2m    &    0.6401        & 0.55\%   & 41m    &  \underline{0.7071}         &\underline{0.10\%}   & 2.6h       \\
PPLS/D-C (200 iter.)     &  0.6256         & 1.29\%    &  25m    &   0.6282        & 2.40\%    &  2.7h    &  0.6845            & 3.30\%      &  11h      \\
NSGAII-TSP               &  0.5941        & 6.26\%    & 40m    &    0.5984       & 7.03\%    &  43m    &    0.6643          &  6.15\%     &    53m    \\ \midrule
DRL-MOA (101 models)     &  0.6257         &  1.28\%   & 7s   &    0.6360       & 1.18\%    & 10s    &      0.6971        & 1.51\%      &  22s      \\
PMOCO (101 pref.)       &   0.6266        & 1.14\%    &  8s   &  0.6349         &  1.35\%   & 13s   &    0.6953          & 1.77\%      &   21s     \\
NHDE-P (101 pref.)      &   0.6288        & 0.79\%    & 4.3m   &   0.6389        & 0.73\%    &  8.3m  &   0.7005           & 1.03\%      &    16m    \\
CDE (101 pref.)      &  \underline{0.6324}         & \underline{0.24\%}    & 18s    &  0.6404        & 0.53\%    &  23s  &      0.7014        &  0.89\%     &  31s      \\ \midrule
EMNH (aug.) &  0.6271         & 1.05\%    & 1.3m   &  0.6408        & 0.44\%    &  4.6m   &    0.7023           & 0.78\%      &   17m \\
PMOCO (101 pref. \& aug.)   &  0.6273         & 1.03\%    & 46s   &   0.6392        & 0.69\%    &  3.1m    &   0.6997           & 1.14\%      &   5.7m     \\
NHDE-P (101 pref. \& aug.)   &  0.6296         &  0.66\%   & 9.8m   &  \underline{0.6429}         &\underline{0.09\%}     &  19m  &     0.7050         &  0.40\%     &   40m     \\
CDE (101 pref. \& aug.) &  \textbf{0.6340}         & \textbf{0.00\%}    & 1.2m   &  \textbf{0.6437}        & \textbf{0.00\%}    &  4.5m   &   \textbf{0.7079}           &  \textbf{0.00\%}     & 6.8m       \\ \midrule     
Method                   & \multicolumn{3}{c|}{Bi-CVRP20}               & \multicolumn{3}{c|}{Bi-CVRP50}               & \multicolumn{3}{c}{Bi-CVRP100}  \\ \midrule
PPLS/D-C (200 iter.)     &  0.3351         & 4.04\%    &  1.2h    &  0.4149         & 3.42\%    &  9.6h    &  0.4083            &  1.87\%     &  37h      \\
NSGAII-CVRP               &   0.3123        & 10.57\%    & 37m    &  0.3631         &  15.48\%   &  38m    &  0.3538            &  14.97\%     &  43m      \\ \midrule
DRL-MOA (101 models)     &     0.3453     & 1.12\%    & 7s   &  0.4270         &  0.61\%   &  20s   &  \textbf{0.4176}          &   \textbf{-0.36\%}    &  40s      \\
PMOCO (101 pref.)       &   0.3467        & 0.72\%    &  8s    &   0.4271        & 0.58\%    &  18s  &   0.4131           &  0.72\%     &   36s     \\
NHDE-P (101 pref.)      &  0.3458        & 0.97\%    & 1.5m   &   0.4248        & 1.12\%    & 3.1m   &  0.4127            &  0.82\%     &   5.3m     \\
CDE (101 pref.)      &  0.3471         & 0.63\%    & 17s   &   0.4279        & 0.47\%    & 31s  &     0.4143         &  0.43\%     &  58s      \\ \midrule
EMNH (aug.) &  0.3471         & 0.60\%    & 33s  &  0.4250        & 1.07\%    &  1.4m  &    0.4146           & 0.36\%      &   3.7m \\
PMOCO (101 pref. \& aug.)   & \underline{0.3481}          & \underline{0.32\%}    & 1m   &   \underline{0.4287}        & \underline{0.21\%}    &  2.1m    &   0.4150           &  0.26\%     &    4.5m    \\
NHDE-P (101 pref. \& aug.)   &    0.3465      &  0.77\%  &  5.1m   &    0.4262       & 0.79\%    & 9.2m   &  0.4139            & 0.53\%      & 21m       \\
CDE (101 pref. \& aug.) &    \textbf{0.3492}       &  \textbf{0.00\%}   & 2.2m   &    \textbf{0.4297}       & \textbf{0.00\%}    & 4.1m    &   \underline{0.4161}        &   \underline{0.00\%}    &   6.8m     \\ \midrule  
Method                   & \multicolumn{3}{c|}{Bi-KP50}               & \multicolumn{3}{c|}{Bi-KP100}               & \multicolumn{3}{c}{Bi-KP200}  \\ \midrule
WS-DP (101 pref.)       &  0.3563         & 0.50\%   & 9.5m    &    0.4531        & 0.77\%   & 1.2h   &  0.3599         & 2.04\%    & 3.7h       \\
PPLS/D-C (200 iter.)     &  0.3528         & 1.48\%    &  18m    &   0.4480        &  1.88\%   & 46m     &  0.3541        & 3.62\%      &  1.4h      \\
NSGAII-KP               &  0.3112         &  13.10\%   &  30m   &    0.3514       &  23.07\%   & 31m     &  0.3511       &  4.44\%     &    33m    \\ \midrule
EMNH              &  0.3561        & 0.56\%    & 17s  &  0.4535        & 0.68\%    &  53s  &   0.3603          & 1.93\%      &   2.3m\\
DRL-MOA (101 models)     &  0.3559         & 0.61\%    &  8s  &  0.4531         & 0.77\%    &  13s   &    0.3601          &  1.99\%     &  1.1m      \\
PMOCO (101 pref.)       &   0.3552        &  0.81\%   &  13s    &  0.4522         & 0.96\%    & 19s   &   0.3595           & 2.15\%      &   50s     \\
NHDE-P (101 pref.)      &   \underline{0.3564}        & \underline{0.47\%}    & 1.1m   &   \underline{0.4541}        & \underline{0.55\%}    &  2.5m  &   \underline{0.3612}           & \underline{1.69\%}      &  5.3m      \\
CDE (101 pref.)      &   \textbf{0.3582}        & \textbf{0.00\%}    &  21s  &   \textbf{0.4571}        &  \textbf{0.00\%}   &  33s &    \textbf{0.3674}          &   \textbf{0.00\%}    &   1.4m     \\ \midrule  
Method                   & \multicolumn{3}{c|}{Tri-TSP20}               & \multicolumn{3}{c|}{Tri-TSP50}               & \multicolumn{3}{c}{Tri-TSP100}  \\ \midrule
WS-LKH (210 pref.)       &  0.4718         &  1.50\%   & 20m    &    0.4493        &  2.50\%   & 3.3h   &   0.5160      & 1.28\%    & 11h       \\
PPLS/D-C (200 iter.)     &  0.4698         & 1.92\%    & 1.3h     &   0.4174        & 9.42\%    &  3.8h    &    0.4376          & 16.28\%      & 13h       \\
NSGAII-TSP               &  0.4216         & 11.98\%    & 2.1h    &   0.4130        &  10.37\%   &  2.3h    &  0.4291            &  17.91\%     &  2.5h      \\ \midrule
DRL-MOA (1035 models)     &   0.4712        & 1.63\%    & 51s   &  0.4396          & 4.60\%    & 1.5s    &    0.4915           & 5.97\%      &  3.1s      \\
PMOCO (10201 pref.)       &  0.4749         & 0.86\%    &  8.9m    &  0.4489         & 2.58\%    & 17m   &    0.5102          &  2.39\%     &  34m      \\
NHDE-P (10201 pref.)      &  0.4764         & 0.54\%    &  53m  &  0.4513         &  2.06\%   & 1.8h   &  0.5118            &  2.09\%     &  4.3h      \\
CDE (10201 pref.)      &  0.4782         & 0.19\%    & 10m   &   0.4531        & 1.71\%    & 19m  &  0.5129            &  1.87\%     &    41m    \\ \midrule
EMNH (aug.)             &  0.4712        & 1.63\%    & 7.1m  &  0.4418        & 4.12\%    &  58m  &   0.4973          & 4.86\%      &   2.4h\\
PMOCO (10201 pref. \& aug.)   &  0.4757          & 0.69\%    & 20m   &   0.4573        & 0.76\%    &   1.1h   &    0.5123          & 1.99\%      &  4.3h      \\
NHDE-P (10201 pref. \& aug.)   &  \underline{0.4772}         & \underline{0.38\%}    &  2.1h  &   \underline{0.4595}        & \underline{0.28\%}    &  6.7h  &     \underline{0.5210}         &  \underline{0.33\%}     &   15.3h     \\
CDE (10201 pref. \& aug.) &   \textbf{0.4791}        & \textbf{0.00\%}    & 26m   &  \textbf{0.4609}         &  \textbf{0.00\%}   &  1.3h   &    \textbf{0.5227}          &  \textbf{0.00\%}     &  4.8h      \\ \midrule 
\end{tabular}
\vspace{-0.1cm}
\end{table*}
\subsection{Ablation and Hyperparameter Studies}
% \begin{minipage}{.4\linewidth} 
To analyze the effect of the HRU strategy, we compare CDE with 4 variants:
% in the training stage: 
CDE without non-local term (CDE w/o non-local), CDE without local term (CDE w/o local), CDE with a constant of $\alpha=1$ (CDE w/o residual) and CDE without local term decay parameter (CDE w/o local decay). Moreover, we compare CDE with  CDE without implicit inference (CDE w/o implicit) and CDE without explicit inference (CDE w/o explicit), to study the $EI^2$ approach.
As seen in Table \ref{t2}. The performance of CDE is impaired when
any of the components is ablated. Besides, both local and non-local terms are critical for model training since their absence leads to the top 2 biggest gaps. The introduction of adaptive parameters $\omega$ and $\alpha$  can effectively control the focus of the Pareto attention model to make the reward function approach HV. \textcolor{black}{We further study the influence of the length of sequence $K$. We compare CDE with larger $K$ (10/16 for 50/100 instance nodes) and with no context information ($K=1$) in Table \ref{t2}. We find that the past information is useful for MOCO since the performance is significantly worse when $K = 1$. Moreover, longer sequence will not get obvious performance improvement. The phenomenon may be based on one hypothesis that when representing a distribution of policies (i.e. sequence modeling), the context allows the attention model to identify which policy generated the actions, enabling better learning and/or improving the training/inference dynamics, and if the sequence length is too long, it will only provide redundant information.}
% We also show the inference time of CDE and CDE w/o LSSA in Figure \ref{time}. CDE w/o LSSA adopts traditional LSS. It is evident that LSSA 
% can improve efficiency, as discussed in Section \ref{44}.

% \begin{minipage}{.6\linewidth} 
\begin{table}[H]  
 \setlength{\tabcolsep}{2pt}
 \setlength{\abovecaptionskip}{0cm} % 调整间距
 \scriptsize
  \caption{Effects of HRU strategy and $\rm{EI^2}$ approach.}
  % \vspace{-0.05cm} 
  \label{t2}
    \centering
  \scalebox{0.9}{\begin{tabular}{l|cc|cc|cc|cc}
   \toprule 
       & \multicolumn{2}{c|}{Bi-TSP50}            & \multicolumn{2}{c|}{Bi-CVRP50}               & \multicolumn{2}{c|}{Bi-KP100}               & \multicolumn{2}{c}{Tri-TSP50} 
                   \\
Method                   & HV $\uparrow$        & Gap $\downarrow$  & HV $\uparrow$       & Gap $\downarrow$  & HV $\uparrow$      & Gap $\downarrow$ & HV $\uparrow$      & Gap $\downarrow$   \\ 
\midrule
CDE w/o non-local&0.6347 &0.86\% &0.4368 &0.25\% &0.4521 &1.07\% &0.4487 &0.95\%\\
CDE w/o local    &0.6251 &2.35\% &0.4239 &3.14\% &0.4416 &3.31\% &0.4326 &4.51\%\\ 
CDE w/o residual &0.6379 &0.36\% &0.4361 &0.36\% &0.4547 &0.53\% &0.4507 &0.53\%\\
CDE w/o local decay &0.6401 &0.03\% &\underline{0.4376} &\underline{0.04\%} &0.4566 &0.11\% &0.4529&0.04\%\\
CDE w larger $K$ &\underline{0.6403} &\underline{0.01\%} &0.4375 &0.53\% &\textbf{0.4572} &\textbf{-0.02\%} &\underline{0.4530}  &\underline{0.02\%}\\
CDE w no context &0.6389 &0.37\% &0.4356 &0.53\% &0.4552 &0.42\% &0.4513 &0.40\%\\
\midrule
CDE w/o implicit&0.6371 &0.48\% &0.4367  &0.27\% &0.4558 &0.28\% &0.4517 &0.31\%\\ 
CDE w/o explicit &0.6363 &0.61\% &0.4348 &0.71\% &0.4549 &0.48\% &0.4511 &0.44\%\\
CDE              &\textbf{0.6404} & \textbf{0.00\%}&\textbf{0.4379}&\textbf{0.00\%}  &\underline{0.4571} &\underline{0.00\%} &\textbf{0.4531}&\textbf{0.00\%}  \\ 
\midrule 
  \end{tabular}}
\end{table}
% \end{minipage}
% \hfill
% \begin{minipage}{.35\linewidth}
% \begin{figure}[H]
% 			\centering
                
%                 % \setlength{\abovecaptionskip}{0.cm}
%                 \includegraphics[width=0.95\textwidth]{Time.pdf}
%                 \caption{The time consumed by CDE and CDE w/o LSSA.} 
%                 % on 4 problems
%                 % strategy improves efficient to a great extent.}
%                 \label{time}
%                 % \setlength{\belowcaptionskip}{-10cm}
                
% \end{figure}
% \end{minipage}
% \end{minipage}
% \begin{minipage}{.8\linewidth} 

% \end{minipage}
% \vspace{-0.7cm}

% \vspace{-0.3cm}
\subsection{Validity of Context Awareness}
\label{validity}
\begin{figure} [H]
\raggedleft
    % \vspace{-0.7cm}
    % \centering
    % \begin{subfigure}[b]{0.49\linewidth}
    %     \centering
    %     \includegraphics[width=\linewidth]{dtlzd6.pdf}
    % \end{subfigure}
    % 238 227
    \subfigure[TCH]{\begin{minipage}[h]{0.085\textwidth}
        \includegraphics[width=1\textwidth]{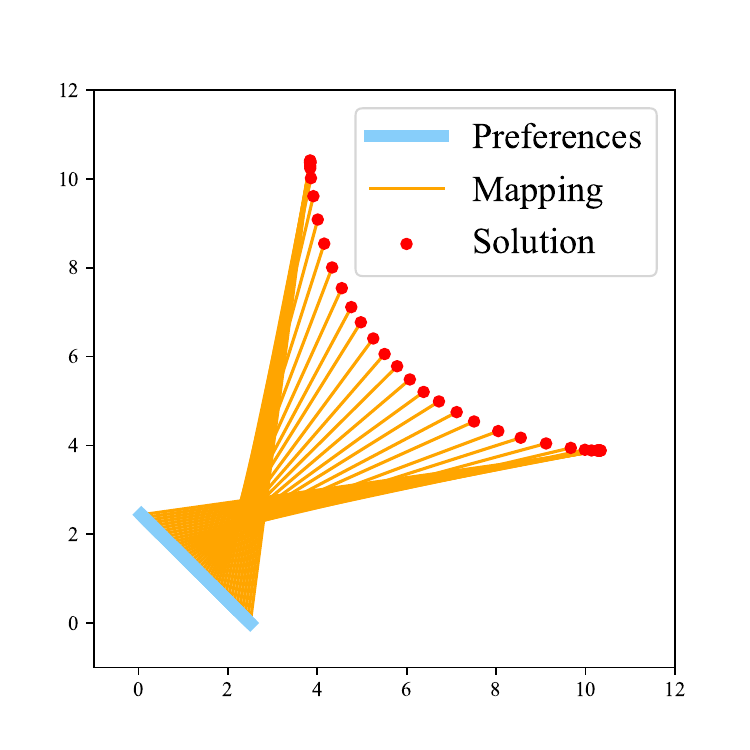}
    \end{minipage}}
    \subfigure[WS]{\begin{minipage}[h]{0.085\textwidth}
        \includegraphics[width=1\textwidth]{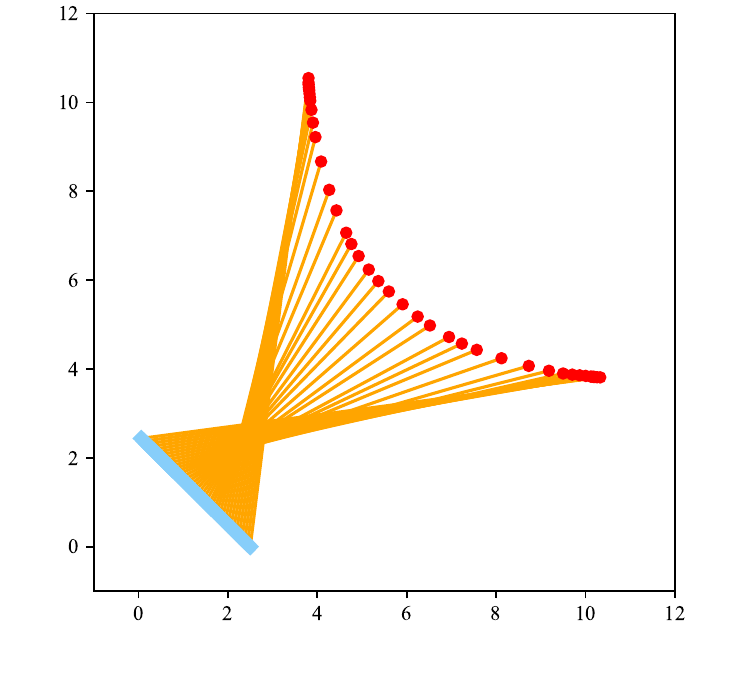}
    \end{minipage}}
    \subfigure[CDE1]{\begin{minipage}[h]{0.085\textwidth}
        \includegraphics[width=1\textwidth]{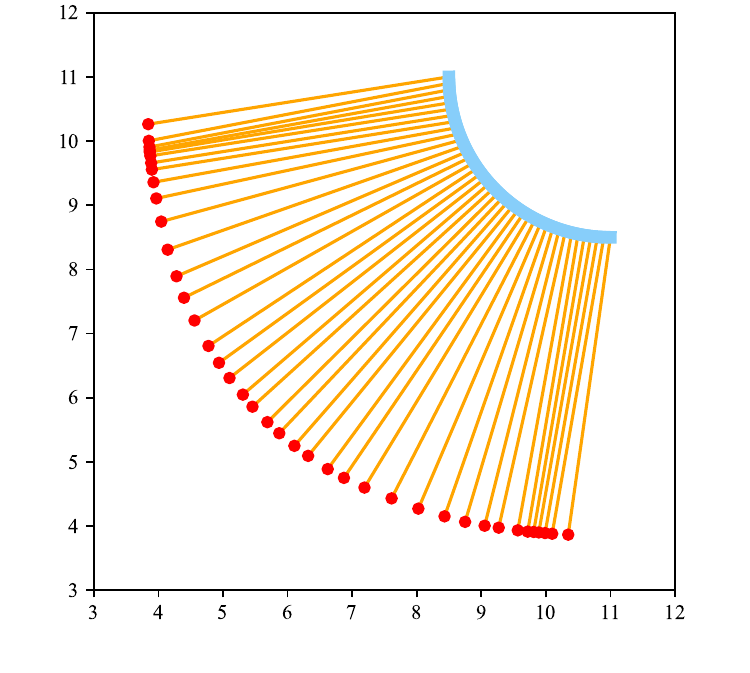}
    \end{minipage}}
    \subfigure[CDE2]{\begin{minipage}[h]{0.085\textwidth}
        \includegraphics[width=1\textwidth]{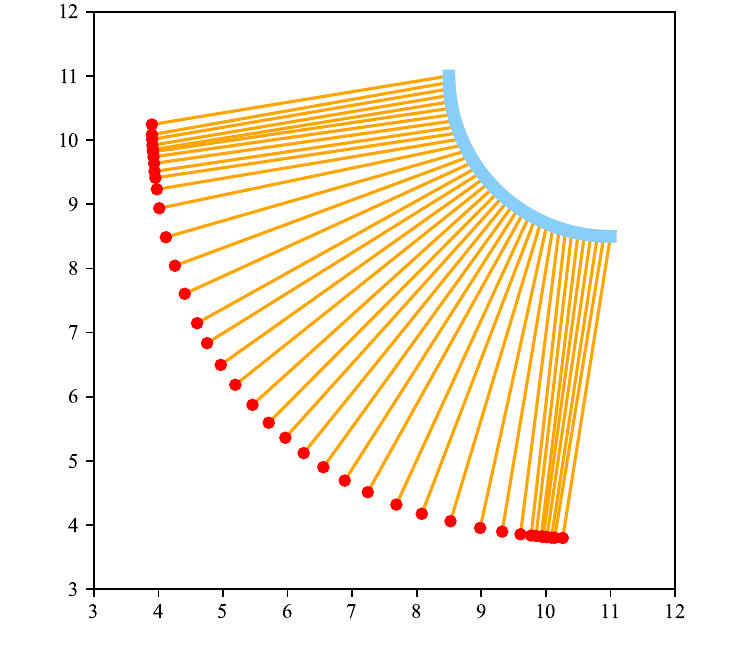}
    \end{minipage}}
        \subfigure[CDE]{\begin{minipage}[h]{0.085\textwidth}
        \includegraphics[width=1\textwidth]{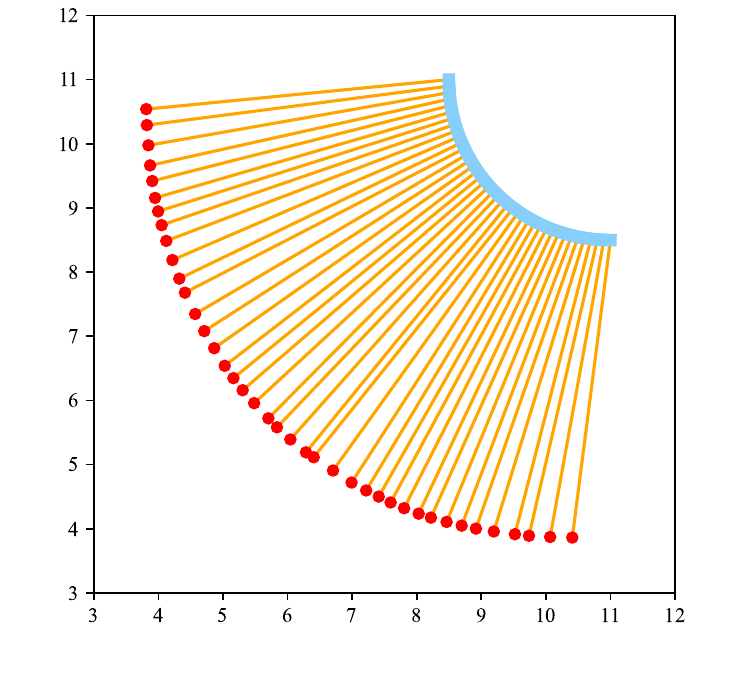}
    \end{minipage}}
    \subfigure[TCH]{\begin{minipage}[h]{0.085\textwidth}
        \includegraphics[width=1\textwidth]{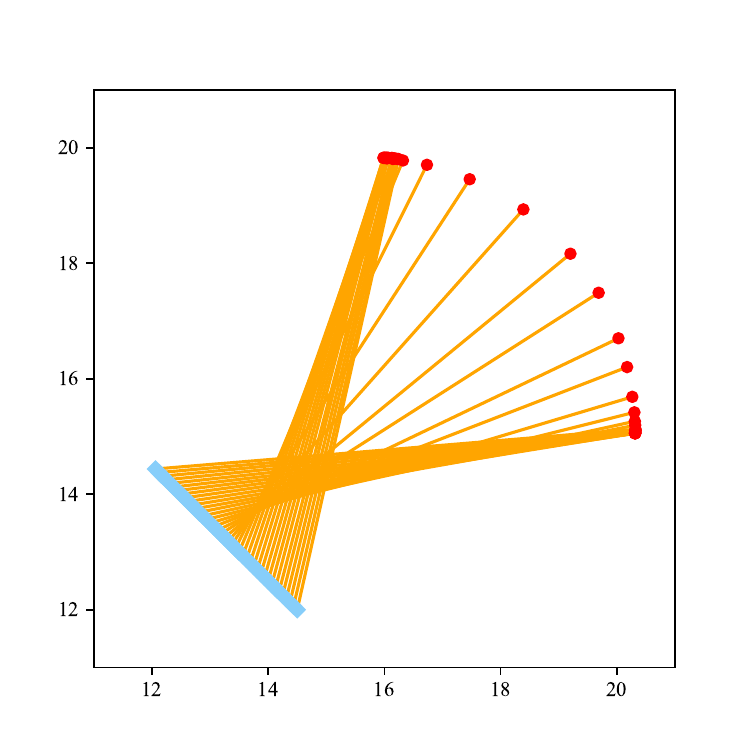}
    \end{minipage}}
    \subfigure[WS]{\begin{minipage}[h]{0.085\textwidth}
        \includegraphics[width=1\textwidth]{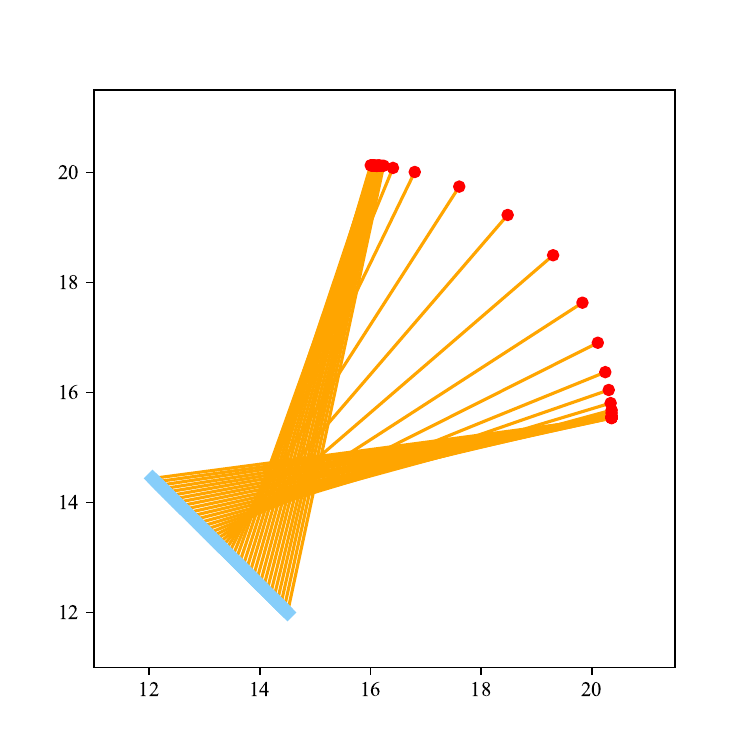}
    \end{minipage}}
    \subfigure[CDE1]{\begin{minipage}[h]{0.085\textwidth}
        \includegraphics[width=1\textwidth]{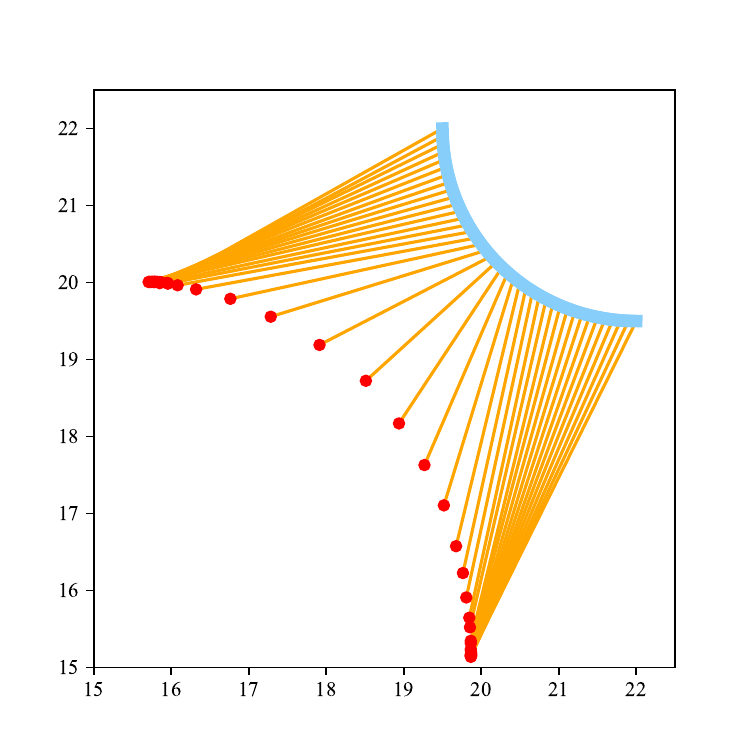}
    \end{minipage}}
    \subfigure[CDE2]{\begin{minipage}[h]{0.085\textwidth}
        \includegraphics[width=1\textwidth]{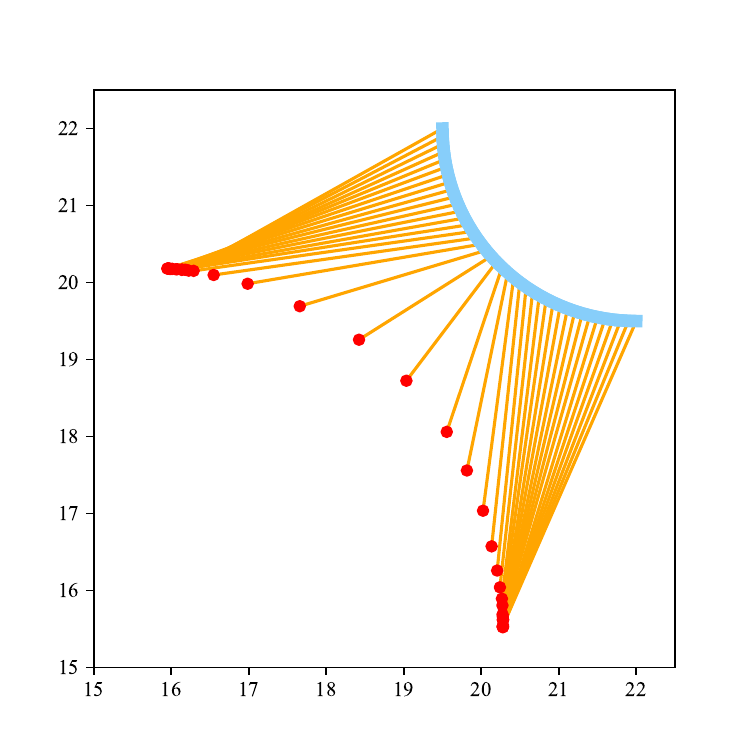}
    \end{minipage}}
        \subfigure[CDE]{\begin{minipage}[h]{0.085\textwidth}
        \includegraphics[width=1\textwidth]{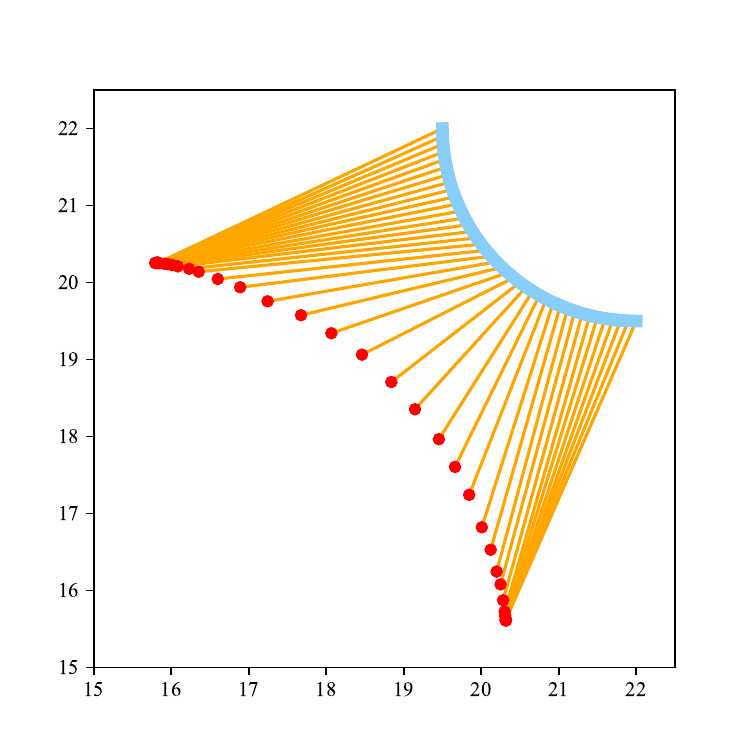}
    \end{minipage}}
    \caption{Visual comparisons on Bi-TSP20 (a-e) and Bi-KP50 (f-j). CDE  has better adaptability than WS- and TCH-based approaches. CDE1 and CDE2 denote the CDE w/o node-level context awareness and solution-level context awareness, respectively. Any ablation of context-aware components leads to a decrease in diversity.}
    \label{mse}
    % \vspace{-0.7cm}
\end{figure}

% \begin{minipage}{.44\linewidth} 
% \raggedleft
\textcolor{black}{As demonstrated earlier,  CDE aims to enhance diversity via context awareness for different subproblems and understand their interconnections. In CDE, the node-level context awareness improves the attention model's understanding of sequence properties and the solution-level context awareness promotes the diversity of PS/PF. The integration of mappings between polar angles (preferences) and solutions approximates the HV. This allows the Pareto attention model to effectively capture these mappings, benefiting from its larger state space compared with traditional heuristic optimization methods. Furthermore, CDE improves previous decomposition-based approaches by building the assumption of unequal importance for all subproblems. As a result, CDE is able to approximate the entire Pareto front and prevent duplicate solutions across different subproblems. Figure \ref{mse} visualizes the mapping of CDE and other decomposition-based methods in the objective space.}

 \textcolor{black}{This visualization supports our idea from two perspectives: 1) Diversity. In terms of the true PF with uniformly distributed solutions (e.g. MOTSP), WS, TCH-based PMOCO, and two CDE variants produce a large number of duplicate solutions at the corner of the PF. In contrast, CDE (Fig. \ref{mse}(e)) realizes uniform mapping on the PF. 
When examining the true PF with non-uniformly distributed solutions (e.g., MOKP), WS-, TCH-based methods, and two variants struggle to find solutions in sparse regions, while CDE (Fig. \ref{mse}(j)) effectively simulates the nonuniform distribution characteristics and successfully locates solutions in such areas. 2) Clarity. The mappings generated by CDE exhibit distinct geometric representations compared with other decomposition-based approaches. These observations lead us to conclude that CDE demonstrates exceptional adaptability due to its context awareness and each component plays a positive role in enhancing diversity. More visualization results are presented in Appendix \ref{more}.}
% \end{minipage}
% \hfill
% \begin{minipage}{.53\linewidth} 

% \end{minipage}

% \end{minipage}

% \vspace{-0.4cm}
\section{Conclusion}
\textcolor{black}{This paper proposes a  
Context-aware Diversity Enhancement method for MOCO, offering a novel context-aware perspective, which achieves node-level context awareness via sequence modeling and solution-level context awareness via hypervolume expectation
maximization.} 
% which improves diversity significantly compared with previous decomposition-based and Markov decisionbased methods
HRU strategy makes the Pareto attention model better utilize the local and non-local information of the PS/PF. Furthermore, the $\rm{EI^2}$ approach provides high-quality solutions with efficient approximate HV while LSSA enhances selection efficiency. Experimental results on three well-known MOCO problems demonstrate the superiority of CDE, particularly in terms of superior decomposition and efficient diversity enhancement. 
% It should be noted that, like other neural heuristics, CDE does not guarantee obtaining the exact PF. 
% For future work, we plan to extend CDE to other constrained MOCO problems and real-world applications. 

\paragraph{Impact Statements.}
This paper presents work whose goal is to advance the field of Multi-Objective Combinatorial Optimization (MOCO). There are many potential societal consequences of our work, none which we feel must be specifically highlighted here.

\bibliography{example_paper}
\bibliographystyle{icml2025}
\newpage
\appendix
\onecolumn

\renewcommand\thesection{\Alph{section}}
\section{Limitations}
\label{limit}
The main limitation of CDE is that the design of our reward function is designed by empirical results. In our experiment, we find that both local and non-local term present different influence in different stages of training. To be specific, the non-local term is difficult to study compared with local term while the non-local term contains more information. Therefore, we introduce two adaptive parameters  to control the  ratio of them according to the state of training. In the  future, we will further study the connection between them to design more suitable reward function with theoretically sound.

\section{Broader Impact}
\label{broader}
In our paper, we propose Context-aware Diversity Enhancement algorithm, The positive impact can be analyzed from two aspects. To begin with, we design a novel reward function, which includes local term and non-local term. It is the first time to introduce global information in the reward function. Therefore, the mapping from preferences to objetcive space in CDE is clearer than the previous methods. Secondly, the introduction of approximate HV can speed up the neural network to extract global information, which has reference significance for the follow-up methods. Lastly, we design a novel inference method compared with previous simple methods.

On the negative side, the reward function without theoretical guarantee shows remarkable improvement in most benchmark problems. However, the performance on real-world engineering design needs to be further tested to ensure safety.

\section{Other Related Work}
\label{related}
\paragraph{Neural Heuristics for MOCO.} 
Decomposition is a mainstream scheme in learning-based methods for multi-objective optimization \cite{lin2022pareto, navon2020learning, lin2019pareto}. 
% An MOCOP can be decomposed into a series of single-objective CO problems and then solved by neural construction methods to approximate the Pareto set. 
% \textcolor{black}{the two sentence are Repetitive.
% I think the second one is better.}
Their basic idea
is to decompose MOCO problems into multiple subproblems according to prescribed weight vectors, and then train a single model or multiple models to solve these subproblems. 
% \textcolor{black}{Better to use people (Zhang et al.) as 
% the subject of verbs like 'propose' or 'train', instead of xx paper }
For example,
\cite{li2020deep} and  \cite{zhang2021modrl} train 
multiple models collaboratively through a transfer learning strategy. 
% Preference-conditioned multi-objective combinatorial optimization (PMOCO) 
\cite{lin2022pareto} train a hypernetwork-based model to generate the decoder parameters conditioned on the preference for MOCO (PMOCO).
Both MDRL \cite{zhang2022meta} and EMNH \cite{chen2023efficient} leverages meta-learning to
train a deep reinforcement learning model that could be fine-tuned for various subproblems. NHDE \cite{chen2023neural} proposes indicator-enhanced DRL with an HGA model, which is the first to introduce the hypervolume into the reward for MOCO to enhance diversity.
% trains a hypernetwork-based model conditioned on the preference., 
% which can generate the decoder parameters according to weight vectors for solving subproblems. 
% Meta-Learning-based DRL (MLDRL) \cite{zhang2022meta} first trains a meta-model and then quickly fine-tunes the meta-model based on 
% weight vectors to solve subproblems
% \textcolor{black}{ based on xx, info density too low?}.
% To our best knowledge, PMOCO and MLDRL are two competitive DRL methods for MOCOPs
% \textcolor{black}{What's the point of this sentence}.

\paragraph{Exact and Heuristic Methods for MOCO.} Exact \cite{florios2014generation} and heuristic \cite{herzel2021approximation} algorithms are two groups of methods to solve MOCO problems in the past decades. The former can find all the Pareto-optimal solutions for only very small-scale problems, while the latter, commonly used in practical applications, can find the approximate Pareto-optimal solutions within a reasonable time. Many multi-objective evolutionary algorithms (MOEAs) are customized for MOCO, including NSGA-II \cite{deb2002fast}, MOEA/D \cite{zhang2007moea}, and PPLS/D-C \cite{shi2022improving}. 
\paragraph{Neural Heuristics for Combinatorial Optimization (CO).} 
In the literature, some end-to-end neural construction methods are developed
for solving SOCO problems. The pioneering works \cite{bello2016neural,nazari2018reinforcement, vinyals2015pointer} trained a pointer network to construct a near-optimal solution for SOCO problems.  Kool et al. ~(\cite{kool2018attention}) proposed
an attention model (AM) based on the transformer architecture. 
Another representative work is policy optimization with multiple optima (POMO) \cite{kwon2020pomo}, which exploits the symmetry of solutions to further improve the performance of end-to-end models. 
Besides, the other line of works, known as neural improvement methods \cite{ma2021learning,lu2019learning}, exploited DRL to assist the iterative improvement process from an initial solution, following a learn-to-improve paradigm.

\section{Hardware Settings}
All the experiments
are conducted with an RTX 4090 GPU and a 1.5GHz AMD EPYC 7742 CPU. 
\section{Aggregation Function}
An aggregated  (or utility) function can map each point in the objective space
into a scalar according to an $m$-dimensional weight vector $\lambda$ with $\|\boldsymbol{\lambda}\|_p=1$ ( $l_p $-norm constraint ) and $\lambda_i \geq 0$. Weighted-Sum (WS) and Weighted-Tchebycheff are commonly used utility functions \cite{miettinen1999nonlinear}. As the simplest representative, WS can be defined by $\min_{\boldsymbol{x} \in \mathcal{X}} f(\boldsymbol{x}|\boldsymbol{\lambda})=\sum^M_{i=1} \lambda_i f_i(\boldsymbol{x})$.
% \textcolor{black}{Is is necessary to introduce agg func
% as an subsection? }
\section{Reference Points and Ideal Points} 
\label{refer}
For a problem, all methods share the same reference point $\boldsymbol{r}$ and ideal point $\boldsymbol{z}$, as shown in Table \ref{synthetic555}.

\begin{table}[!h]  
    %  \scriptsize
    
    \footnotesize
        % \tiny\scriptsize\footnotesize\small\normalsize\large\Large\LARGE\huge\Huge
 % \tabcolsep0.01in
 \renewcommand{\arraystretch}{1.3}
  \caption{Description of synthetic benchmarks we used in this work.}
  \label{synthetic555}
    \centering
  \begin{tabular}{lllll}
    \toprule
    Problem&Size&reference point ($\boldsymbol{r}$)&ideal point ($\boldsymbol{z}$)\\
    \midrule
    Bi-TSP &20&(20, 20) &(0, 0)\\
    &50&(35, 35)&(0, 0)\\
    &100&(65, 65)&(0, 0)\\
    \midrule
    Bi-CVRP &20&(15, 3)&(0, 0)\\
    &50&(40, 3)&(0, 0)\\
    &100&(60, 3)&(0, 0)\\
    \midrule
    Bi-KP &50&(5, 5)&(30, 30)\\
    &100&(20, 20)&(50, 50)\\
    &200&(30, 30)&(75, 75)\\
    \midrule
    Tri-TSP &20&(20, 20, 20)&(0, 0, 0)\\
    &50&(35, 35, 35)&(0, 0, 0)\\
    &100&(65, 65, 65)&(0, 0, 0)\\
    \bottomrule
    
  \end{tabular}
\end{table}

\section{Extension of Lemma \ref{lemma1} to Infinite Set} 
\label{extension}
Lemma \ref{lemma1}  with a finite number of objective vectors can be extended to an infinite set, which can also be expressed as an expectation \cite{zhang2023hypervolume}:
\begin{figure}[!htbp]
			\centering

                \includegraphics[width=0.3\textwidth]{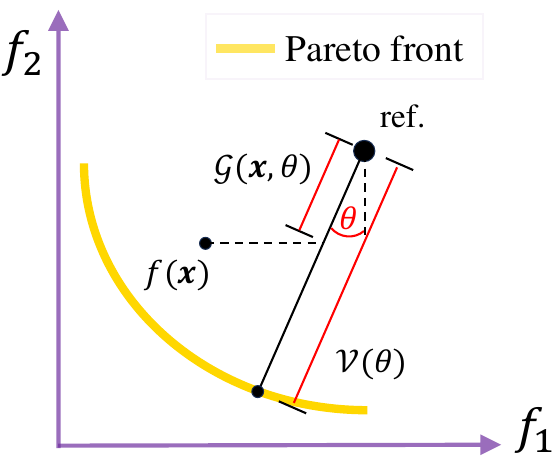}
                \caption{Pareto front hypervolume calculation in the polar coordinate. $\mathcal{G}(\boldsymbol{x},\theta)$ is the distance from the reference point to the Pareto front along angle $\theta$. $\mathcal{V}(\theta)$ is the projected distance at angle $\theta$.}
                \label{fig1}
                % \vspace{-1.1cm}
\end{figure}
\begin{eqnarray} 
            \begin{aligned}
            \mathcal{HV}_{\boldsymbol{r}}(Y) = \frac{\Phi}{m 2^m} \mathbb{E}_{\theta}[(\mathcal{V}_{\mathcal{X}}(\theta)^m)] 
              \end{aligned},      
            \end{eqnarray}
% \vspace{-0.3cm}
where  $\mathcal{V}_{\mathcal{X}}(\theta)$ denotes the Euclidean distance between the reference point  and the PF with the polar angle $\theta$ (see
Figure \ref{fig1}). Moreover, $\mathcal{V}_{\mathcal{X}}(\theta)$ can be precisely evaluated using the following equation:
\begin{eqnarray} 
            \begin{aligned}
            \label{easy}
   \mathcal{V}_{\mathcal{X}}(\theta)&=\max_{\boldsymbol{x} \in \mathcal{X}}\mathcal{G}^{mtch}(f(\boldsymbol{x}),\theta)=\max_{\boldsymbol{x} \in \mathcal{X}}\min_{i \in \{1,...,m\}}\{(r_i-f_i(x_i))/\lambda_i(\theta)\},
              \end{aligned}           
            \end{eqnarray}
where $\mathcal{G}^{mtch}(f(\boldsymbol{x}),\theta)$ represents the projected distance of an objective vector $f(\boldsymbol{x})$ with polar angle $\theta$ (see
Figure \ref{fig1}). Moreover, the optimal solution $\boldsymbol{x}$ of $\mathcal{V}_{\mathcal{X}}(\theta)$ has the following properties: 1) satisfies $\frac{r_1-y_1(\boldsymbol{x})}{\lambda_1(\theta)} =...=\frac{r_m-f_m(\boldsymbol{x})}{\lambda_m(\theta)}$; 2)
is Pareto optimal. Directly optimizing  $\mathcal{HV}_r$  in Eq. \ref{hvv} is challenging (time-consuming).
However,  \cite{zhang2023hypervolume} provides an easy-to-compute form of
the HV, denoted by $\widetilde{HV_r}(\beta)$:
\begin{eqnarray} 
            \begin{aligned}
            \label{comput}
   &\widetilde{\mathcal{HV}_r}(\beta, \boldsymbol{\theta}^{P'}, s)=\frac{\Phi}{m 2^m} \mathbb{E}_{ \boldsymbol{\theta}^{P'}}[(\mathcal{V}_{\beta} (\theta^i,s))^m] \\
   &\mathcal{V}_{\beta}(\theta,s) = \max(\mathcal{G}^{mtch}(f(\Psi_{\beta}(\theta, s)),\theta),0))
              \end{aligned},           
            \end{eqnarray}
where $\boldsymbol{\theta}^{P'}=\{\theta^1,...\theta^{P'}\}$ denotes the polar angles for all the subproblems in the polar
angles pool, $\mathcal{V}_{\beta}(\theta,s)$ is the projected distance in Eq. \ref{easy} ($0 \leq \mathcal{V}_{\beta}(\theta,s) \leq \mathcal{V}_{\mathcal{X}}(\theta)$ if there is an optimal solution in $\mathcal{X}$. When
the Pareto attention model is well-trained,
% (i.e. learn the whole characteristics of Pareto front/set)
$\mathcal{V}_{\beta}(\theta,s) \rightarrow \mathcal{V}_{\mathcal{X}}(\theta)$, $\forall \theta \in \rm{Unif}(\Theta)$ and thus $\widetilde{\mathcal{HV}_r}(\beta) \rightarrow \mathcal{HV}_r(Y)$.
\section{Details of Local Subset Selection Acceleration}
\label{detailslssa}
\subsection{Local Subset Selection}
For an optimization problem $S^*=\mathrm{arg} \min_{S^ \in \Omega}f(S)$, where $S$ denotes a subset, $\Omega$ denotes the search space and $f(\cdot)$ denotes quantization function of subset quality, a general
procedure of a local selection algorithm can be summarized as
follows: 
\begin{enumerate}

\item Select an initial subset $S$;
\item Define the neighborhood $\mathcal{N}(S) \subset \Omega$;
\item Move to a neighbor $S$ $\leftarrow$ $S' \in \mathcal{N}(S) $ if $f(S') < f(S)$; 
\item Go to step 2 if the termination condition is not met;
otherwise, output $S$.
\end{enumerate}
\begin{definition}[Neighborhood]
In our case, a solution $S$ in a local selection
is actually a set that contains $\mu$ solutions of the multi-objective problems, the
search space $\Omega=\{ S|S \subset Q, \left| S \right| = \mu \}$ contains all the subsets
of the given population $Q$ with size $\mu$. It is clear that $|\Omega|=C^{\mu}_{\mu+\lambda}$. Define the distance between $S$ and $S'$ as the size of their
symmetric difference:
\begin{eqnarray} 
\label{diff}
            \begin{aligned}
   dis(S,S')=\left|\{ \boldsymbol{x}|\boldsymbol{x} \in S, \boldsymbol{x} \notin S'\} \cup \{ \boldsymbol{x} \in S', \boldsymbol{x} \notin S \}\right|
              \end{aligned}.           
            \end{eqnarray}
Then, the neighborhood of $S$ can be defined as follows:
\begin{eqnarray} 
\label{diff1}
            \begin{aligned}
   \mathcal{N}(S)=\{S'|dis(S,S') \leq \epsilon, S' \in \Omega \}
              \end{aligned},           
            \end{eqnarray}
\end{definition}
where $\epsilon$ is a threshold that defines the range of the neighborhood of S. It is clear that the minimum value of $\epsilon$ is 2, and in
such case the size of the neighborhood is $\left| \mathcal{N}(S) \right|=\mu \times \lambda$. In this article, we only consider the situation when $\epsilon=2$. In such
case, the neighborhood can be also formulated as follows:
\begin{eqnarray} 
\label{diff2}
            \begin{aligned}
   \mathcal{N}(S)=\{S \backslash \{\boldsymbol{x}\} \cup \{\boldsymbol{x}'\}| \boldsymbol{x} \in S, \boldsymbol{x}' \in P \backslash S \}
              \end{aligned}.           
            \end{eqnarray}
In implementation, we can enumerate all the neighbor solutions of $S$ by replacing an element with another one from the
set $Q\backslash S$.
\begin{algorithm} [!htbp]
	%\footnotesize 
	% \LinesNumbered
        
	%\SetAlgoNoLine
      % \centering
	\caption{Local Subset Selection} \label{psl}  
    \begin{algorithmic}[1]
    % \SetKwData{Q}{Q}
	%\SetKwData{y}{y}
	% \SetKwInOut{Input}{input}\SetKwInOut{Output}{output}
% \begin{algorithmic}[1]
    \STATE {\bfseries Input:} the given population $Q_1$, quantization function of subset quality $f$
     % \Output{the optimal subset $Q^*$}
     \STATE Generate an initial subset $Q^* \subset Q_1$  
     % \BlankLine  
   \STATE  Set $Q_2=Q_1 \backslash Q^*$ \\
    \WHILE{True}
       \STATE  Set $x,y=\textrm{arg} \min_{x \in Q^*, y \in Q_2} f(Q^* \backslash \{x\} \cup \{y\})$ 
        \IF{$f(Q^* \backslash \{x\} \cup \{y\})<f(Q^*)$}
         \STATE $Q^*=Q^* \backslash \{x\}  \cup \{y\}$ 
         \STATE $Q_2=Q_2 \backslash \{y\}  \cup \{x\}$
        \ELSE
        \STATE Output $Q^*$ and stop
        \ENDIF
    \ENDWHILE
         \STATE {\bfseries Output:}  the optimal subset $Q^*$
    \end{algorithmic}
\end{algorithm}

Algorithm \ref{psl} presents the proposed subset selection based
on a local search. In line 2, the initial subset is generated by
using the parent solutions; in line 5, the best neighbor of $P^*$
is found by enumerating all of its neighbors; in lines 6–8,
the subset moves to its neighbor if a better one is found; the
algorithm terminates in line 10 when no better neighbor could
be found, and a local optimal subset is guaranteed.

\subsection{Advantages of Potential Energy Function}
The potential energy model has been widely used in
discretizing manifolds \cite{hardin2004discretizing,borodachov2019discrete}, Recently, several traditional optimization researchers tried to adopt similar ideas to improve the diversity of solutions \cite{falcon2020exploiting,gomez2017hyper} and/or to generate well-distributed
reference points \cite{blank2020generating}, since it has the following properties.
\begin{enumerate}

\item High Sensitivity: It is highly sensitive to data distribution. A local unevenness of solutions will intensely
contribute to the overall metric value, so it could provide
strong guidance for the spacing of solutions, providing
high selection pressure;
\item Low Computing Complexity: Compared with the HV
metric, the potential energy of a system is easy to calculate. An accurate value can be calculated with a
time complexity of $O(n_1^2m)$. Besides, it satisfies the
following additional rule, which favors the calculation:
\begin{align}\left\{\begin{aligned}
&E_{x \notin Q}(Q \cup \{\boldsymbol{x}\}) = E(Q)+2\sum_{\boldsymbol{y} \in Q}U(\boldsymbol{y},\boldsymbol{x})\\
        &E_{x \in Q}(Q \backslash \{\boldsymbol{x}\}) = E(Q)-2\sum_{\boldsymbol{y} \in Q, \boldsymbol{y} \neq \boldsymbol{x}}U(\boldsymbol{y},\boldsymbol{x})\\
\end{aligned}.\right.\end{align}

\item Geometric Insensitive: The minimal energy configuration is shape-insensitive, which means uniformly distributed
solutions can be obtained regardless of the shape of the
optimal PF by minimizing the potential energy.
\end{enumerate}
For these reasons, our algorithm uses the potential energy
function as the subset selection objective, i.e., $f(S) = E(S)$. Moreover, We recommend using $c = 2m$
in this article via ablation study.

\subsection{Proofs of the Difference of Contributions Between Solutions}
\label{proof1}
$Q_1$ contains currently selected solutions and $Q_2$ contains
the unselected ones. Then, the contribution of a selected one $e1(\boldsymbol{x})$ and an unselected one $e2(\boldsymbol{x})$ according to the systems $Q_1$
can be defined as follows:
\begin{align}\left\{\begin{aligned}
    \label{gg}
        &e1(\boldsymbol{x})=\sum_{\boldsymbol{y} \in Q_1 \backslash \boldsymbol{x}} U(\boldsymbol{x},\boldsymbol{y})\\
        &e2(\boldsymbol{x})=\sum_{\boldsymbol{y} \in Q_1} U(\boldsymbol{x},\boldsymbol{y})\\
\end{aligned}.\right.\end{align}

Also, The potential energy of a system Q is defined as follows:
\begin{eqnarray} 
\label{Energya}
            \begin{aligned}
   &E(Q)=\sum_{\boldsymbol{x}^i \in Q} \sum_{\boldsymbol{x}^j \in P \backslash{\boldsymbol{x}^i}} U(\boldsymbol{x}^i,\boldsymbol{x}^j), \\ &U(\boldsymbol{x}^i,\boldsymbol{x}^j)=\frac{1}{\|{\boldsymbol{x}^i-\boldsymbol{x}^j}\|^c}
              \end{aligned},    
\end{eqnarray}

Thus, induced from Eq. \ref{Energya} and Eq. \ref{gg}, the following equality
always holds:
\begin{align}\begin{aligned}
    \label{gg2}
        &E(P \backslash \{\boldsymbol{x}\}\cup\{\boldsymbol{y}\}) = E(P)-e1(\boldsymbol{x}) + e2(\boldsymbol{y})-U(\boldsymbol{x},\boldsymbol{y}).
\end{aligned}\end{align}
% It is clear that when evaluating the neighbor of P by replacing
% $\boldsymbol{x}$ by $\boldsymbol{y}$, it is not necessary to calculate the energy of the new
% system. Instead, only the difference between the energy values
% of the two systems is needed:
Consider one of $Q_1$'s neighbors $Q_1 \backslash \{\boldsymbol{x}\} \cup \{\boldsymbol{y}\}$, where $\boldsymbol{x} \in Q_1$ and $\boldsymbol{y} \in Q_2$. We can have the following proof for the difference between the energy of  $Q_1$ and $Q_1 \backslash \{\boldsymbol{x}\} \cup \{\boldsymbol{y}\}$:
\begin{align}\begin{aligned}
    \label{prove}
    \Delta E(Q_1) &= E(Q_1) - E(Q_1 \backslash \{\boldsymbol{x}\} \cup \{\boldsymbol{y}\}) \\
         &= E(Q_1) - 
        \sum_{\boldsymbol{x}' \in Q_1 \backslash \{\boldsymbol{x}\} \cup \{\boldsymbol{y}\}} \sum_{\boldsymbol{x}'' \in Q_1 \backslash \{\boldsymbol{x}, \boldsymbol{x}'\} \cup \{\boldsymbol{y}\}} U(\boldsymbol{x}',\boldsymbol{x}'')\\
        &= E(Q_1) - \left(\sum_{\boldsymbol{x}' \in Q_1} \sum_{x'' \in Q_1 \backslash \{\boldsymbol{x}'\}}U(\boldsymbol{x}',\boldsymbol{x}'')-\sum_{x' \in Q_1 \backslash \{\boldsymbol{x}\} }U(\boldsymbol{x},\boldsymbol{x}') + \sum_{\boldsymbol{x}' \in Q_1}U(\boldsymbol{y},\boldsymbol{x}')-U(\boldsymbol{x},\boldsymbol{y})\right) \\
        &=E(Q_1) - \left(E(Q_1)-e1(\boldsymbol{x})+e2(\boldsymbol{y})-U(\boldsymbol{x},\boldsymbol{y})\right) \\
        &=e1(\boldsymbol{x})-e2(\boldsymbol{y})+U(\boldsymbol{x},\boldsymbol{y}).
\end{aligned}\end{align}
It is clear that when evaluating the neighbor of $Q_1$ by replacing $\boldsymbol{x}$ by $\boldsymbol{y}$, it is not necessary to calculate the energy of the new
system. Instead, only the difference between the energy values
of the two systems is needed. \hfill $\square$

Similarly, we can calculate the difference between the
contributions of selected and unselected solutions. 

Considering a solution $\boldsymbol{x}' \in Q_1$, if $\boldsymbol{x}' \neq \boldsymbol{x}$, we can have following proof:
\begin{align}\begin{aligned}
    \label{prove2}
        \Delta e1(\boldsymbol{x}')&= \sum_{\boldsymbol{x}'' \in Q_1 \backslash \{\boldsymbol{x'}\}}U(\boldsymbol{x}',\boldsymbol{x}'') - \sum_{\boldsymbol{x}'' \in Q_1 \backslash \{\boldsymbol{x}',\boldsymbol{x}\} \cup \{\boldsymbol{y}\}}U(\boldsymbol{x}',\boldsymbol{x}'') \\
        &=\left(\sum_{\boldsymbol{x}'' \in Q_1 \backslash \{\boldsymbol{x}',\boldsymbol{x}\}}U(\boldsymbol{x}',\boldsymbol{x}'') + U(\boldsymbol{x}',\boldsymbol{x})\right) - \left(\sum_{\boldsymbol{x}'' \in Q_1 \backslash \{\boldsymbol{x}',\boldsymbol{x}\}}U(\boldsymbol{x}',\boldsymbol{x}'')+U(\boldsymbol{x'},\boldsymbol{y})\right)\\
        &=U(\boldsymbol{x}',\boldsymbol{x})-U(\boldsymbol{x'},\boldsymbol{y}).
\end{aligned}\end{align}
If $\boldsymbol{x}' = \boldsymbol{x}$:

\begin{align}\begin{aligned}
    \label{prove3}
        \Delta e1(\boldsymbol{x}')&= \sum_{\boldsymbol{x}'' \in Q_1 \backslash \{\boldsymbol{x'}\}}U(\boldsymbol{x}',\boldsymbol{x}'') - \sum_{\boldsymbol{x}'' \in Q_1 \backslash \{\boldsymbol{x}'\} \cup \{\boldsymbol{y}\}}U(\boldsymbol{x}',\boldsymbol{x}'') \\
        &=\sum_{\boldsymbol{x}'' \in Q_1 \backslash \{\boldsymbol{x'}\}}U(\boldsymbol{x}',\boldsymbol{x}'') - \left(\sum_{\boldsymbol{x}'' \in Q_1 \backslash \{\boldsymbol{x}'\}}U(\boldsymbol{x}',\boldsymbol{x}'')+U(\boldsymbol{x'},\boldsymbol{y})\right)\\
        &=-U(\boldsymbol{x'},\boldsymbol{y}).
\end{aligned}\end{align}
Considering a solution $\boldsymbol{y}' \in Q_2$, if $\boldsymbol{y}' \neq \boldsymbol{y}$, we can have following proof:
\begin{align}\begin{aligned}
    \label{prove4}
        \Delta e2(\boldsymbol{y}')&= \sum_{\boldsymbol{x}'' \in Q_1 }U(\boldsymbol{y}',\boldsymbol{x}'') - \sum_{\boldsymbol{x}'' \in Q_1 \backslash \{\boldsymbol{x}\} \cup \{\boldsymbol{y}\}}U(\boldsymbol{y}',\boldsymbol{x}'') \\
        &=\sum_{\boldsymbol{x}'' \in Q_1 }U(\boldsymbol{y}',\boldsymbol{x}'')-\left( \sum_{\boldsymbol{x}'' \in Q_1 }U(\boldsymbol{y}',\boldsymbol{x}'') - U(\boldsymbol{y}',\boldsymbol{x})+U(\boldsymbol{y}',\boldsymbol{y}) \right)\\
        &=U(\boldsymbol{y}',\boldsymbol{x})-U(\boldsymbol{y}',\boldsymbol{y}).
\end{aligned}\end{align}

If $\boldsymbol{y}' = \boldsymbol{y}$:
\begin{align}\begin{aligned}
    \label{prove5}
        \Delta e2(\boldsymbol{y}')&= \sum_{\boldsymbol{x}'' \in Q_1 }U(\boldsymbol{y}',\boldsymbol{x}'') - \sum_{\boldsymbol{x}'' \in Q_1 \backslash \{\boldsymbol{x}\} \}}U(\boldsymbol{y}',\boldsymbol{x}'') \\
        &=\sum_{\boldsymbol{x}'' \in Q_1 }U(\boldsymbol{y}',\boldsymbol{x}'')-\left( \sum_{\boldsymbol{x}'' \in Q_1 }U(\boldsymbol{y}',\boldsymbol{x}'') - U(\boldsymbol{y}',\boldsymbol{x}) \right)\\
        &=U(\boldsymbol{y}',\boldsymbol{x}).
\end{aligned}\end{align}

In summary:
\begin{eqnarray}
\begin{aligned}
    \label{summary}
        &\Delta e1(\boldsymbol{x}')=
\begin{cases}
U(\boldsymbol{x}',\boldsymbol{x})-U(\boldsymbol{x'},\boldsymbol{y}),& \boldsymbol{x}' \in Q_1 \backslash \{\boldsymbol{x}\}\\
-U(\boldsymbol{x'},\boldsymbol{y}), & \boldsymbol{x}'=\boldsymbol{x}
\end{cases}
\\
&\Delta e2(\boldsymbol{y}')=
\begin{cases}
U(\boldsymbol{y}',\boldsymbol{x})-U(\boldsymbol{y}',\boldsymbol{y}),& \boldsymbol{y}' \in Q_2 \backslash \{\boldsymbol{y}\}\\
U(\boldsymbol{y}',\boldsymbol{x}), & \boldsymbol{y}'=\boldsymbol{y}
\end{cases}
\end{aligned}.\end{eqnarray}
\hfill $\square$

\section{Model Architecture}
\label{Model}
The core architecture of the base model is similar with POMO, composed of an encoder and a decoder. For node features $\boldsymbol{x}_1,...,\boldsymbol{x}_n$ and previous solutions $\mathcal{F}$, the encoder first computes
initial node embeddings $\boldsymbol{h}_1^{(0)},...\boldsymbol{h}_n^{(0)} \in \mathcal{R}^d(d=128)$ and previous solution information  by a linear projection.  The final node embeddings $\boldsymbol{h}_1^{(L)},...\boldsymbol{h}_n^{(L)}$ are further computed by $L=6$ attention layers. Each attention layer is
composed of a multi-head attention (MHA) with $M=8$ heads and fully connected feed-forward
 sublayer. Each sublayer adds a skip-connection and batch normalization, as follows:
\begin{align}\begin{aligned}
    \label{g2g}
        &\boldsymbol{h}_i^{(l)}=\rm{BN}(\hat{\boldsymbol{h}_i}+\rm{FF}(\hat{\boldsymbol{h}_i})).
\end{aligned}\end{align}
The decoder sequentially chooses a node according to a probability distribution produced by the
node embeddings to construct a solution. The total decoding step $T$ is determined by the specific
problem. The HV network is employed to tackle the preference $\theta$ for the corresponding subproblem. Specifically, according to the given $\theta$, the HV network generates the decoder parameters of the MHA model $\beta$, which is an encoder-decoder-styled architecture, i.e., $\beta(\theta, s)=[\beta_{en}(s),\beta_{de}(\boldsymbol{\pi})]$. The HV network adopts a simple MLP model with two 256-dimensional
hidden layers and ReLu activation. The MLP first maps an input with $M + 2$ dimensions to a hidden
embedding $g(\theta)$, which is then used to generate the decoder parameters by linear projection.
At step $t$ in the decoding procedure, the glimpse $\boldsymbol{q}_c$ of the context embedding $\boldsymbol{g}_c$  is computed by the MHA layer. 
Then, the compatibility $u$ is calculated as follows,
\begin{eqnarray}
\begin{aligned}
    \label{gggg}
        &u_i=
\begin{cases}
-\infty,& $\rm{node}$ $ i$ $ \rm{is}$ $ \rm{masked}$ \\
C \cdot tanh(\frac{\boldsymbol{q}_c^T(W^K\boldsymbol{h}_u^{(L))}}{\sqrt{d/Y}}),& $\rm{otherwise}$
\end{cases}\end{aligned}.\end{eqnarray}
where $C$ is set to 10 \cite{kool2018attention}. Finally, softmax is employed to calculate the selection probability
distribution $Prob_{\beta}(\boldsymbol{\pi}|s,\theta^i,\mathcal{F}_i^j)$ for nodes, i.e., $Prob_{\beta}(\pi_t|\boldsymbol{\pi}_{1:t-1},s,\theta^i,\mathcal{F}_i^j)$ = Softmax($u$).
% the $query$ $\boldsymbol{g}_c \in \mathcal{R}^d$  is computed by an MHA layer
\section{Node Features and Context Embedding}
\label{Embedding}
The input dimensions of the node features vary with different problems. The inputs of the $m$-objective
TSP are $T$ nodes with $2m$-dimensional features. The inputs of Bi-CVRP are $T$ customer nodes with
3-dimensional features and a depot node with 2-dimensional features. The inputs of Bi-KP are $T$
nodes with 3-dimensional features. At step $t$ in the decoder, a context embedding $\boldsymbol{g}_c$ is used to calculate the probability of node
selection. For MOTSP, $\boldsymbol{g}_c$  is defined as the embedding of the first node $\boldsymbol{h}$$_{\boldsymbol{\pi}_1}$, and the nearest $K$ nodes embeddings \{$\boldsymbol{h}$$_{\boldsymbol{\pi}_{t-K}}$,...,$\boldsymbol{h}$$_{\boldsymbol{\pi}_{t-1}}$\}. For MOCVRP, $\boldsymbol{g}_c$ is defined as 
% the concatenation of the graph embedding $\bar{\boldsymbol{h}}$, 
the nearest $K$ nodes embeddings \{$\boldsymbol{h}$$_{\boldsymbol{\pi}_{t-K}}$,...,$\boldsymbol{h}$$_{\boldsymbol{\pi}_{t-1}}$\}, and the remaining vehicle capacity. For MOKP, $\boldsymbol{g}_c$ is defined as 
% the concatenation of the graph embedding $\bar{\boldsymbol{h}}$, 
the nearest $K$ nodes embeddings \{$\boldsymbol{h}$$_{\boldsymbol{\pi}_{t-K}}$,...,$\boldsymbol{h}$$_{\boldsymbol{\pi}_{t-1}}$\} and the remaining knapsack capacity.

A masking mechanism is adopted in each decoding step to ensure the  feasibility of solutions. For MOTSP,
the visited nodes are masked. For MOCVRP (MOKP), besides the visited nodes, those with a demand
(weight) larger than the remaining vehicle (knapsack) capacity are also masked.

\section{Inference Process of CDE}
In terms of the inference process of CDE, we have adopted  explicit and implicit dual inference (line 11-12) in \ref{infgadpsl} to generate solutions and applied local subset selection acceleration strategy to select elite solutions. 
\begin{algorithm} [!t]
    % \LinesNumbered
	\caption{The inference process of CDE} \label{infgadpsl} 
\begin{algorithmic}[1]
    % \SetKwData{Q}{Q}
    % \SetKwInOut{Input}{input}\SetKwInOut{Output}{output}
     \STATE {\bfseries Input:}  preference distribution $\Theta$,instance distribution $\mathcal{S}$, number of uniformly distributed preferences (polar angles) $P$,  batch size $B$, instance size $N$,  model parameter $\beta$, length of sequence $K$
    
    % \BlankLine  
    % \STATE Initialize the model parameters $\beta$   
    \STATE $Pref$ $\leftarrow$ $P$ uniformly  distributed preferences are generated according to  $\Theta$.
    % \FOR{$e=1$ {\bfseries to} $E$}
   \STATE $s_i$ $\sim$ {\bfseries SampleInstance} ($\mathcal{S}$) \quad $\forall i$ $ \in \{1,...B\}$ 
    \STATE Initialize $\mathcal{F}^j_i$ $\leftarrow$ $\emptyset$ \quad $\forall i
    \forall j \in \{1,...N\}$ 
   \STATE Initialize $\boldsymbol{\theta}^0$ $\leftarrow$ $\emptyset$ 
    \STATE Initialize $\boldsymbol{Y}$ $\leftarrow$ $\emptyset$ 
    \FOR{$p=1$ {\bfseries to} $P$}
        % \STATE $\theta$ $\sim$  {\bfseries  SamplePreference}($\Theta$) 
       \STATE $\boldsymbol{\theta}^{p}$ $\leftarrow$ $\boldsymbol{\theta}^{p-1} \cup {Pref}_p$  
       \STATE $\pi^j_{i}$ $\sim$ {\bfseries  SampleSolution} ($Prob_{\beta(\theta)}(\cdot|s_i,\theta),\mathcal{F}^j_i$, $K$)
       \STATE $\mathcal{V}_{\beta}(\theta,s)$,$\widetilde{\mathcal{HV}_r}(\beta, \boldsymbol{\theta}^{p'}, s)$ $\leftarrow$ Calculate the projection distance and approximate HV by Eq. \ref{comput} $\forall i,j$ 
        
       \STATE $\alpha^j_i$ $\leftarrow$ Calculate expected HV improvement by Eq. \ref{ehvi} \quad $\forall i,j$ 
       \STATE $R^j_{i,1}$ $\leftarrow$ Calculate the reward only by the local term of Eq. \ref{reward} \quad $\forall i,j$ //Explicit inference 
       \STATE  $R^j_{i,2}$ $\leftarrow$ Calculate the reward by both the local and the non-local term of Eq. \ref{reward} according to parameter $\alpha^j_i$ \quad $\forall i,j$ //Implicit  inference 
      \STATE  $\pi_{i,1}$,$\pi_{i,2}$ $\leftarrow$ {\bfseries  Argmax}($R^j_{i,1}$), {\bfseries  Argmax}($R^j_{i,2}$) $\forall j$ 
        % \STATE $b_i$ $\leftarrow$ $\frac{1}{N} \sum^n_{j=1}$($R^j_i$) \quad $\forall i$
        % \STATE $\nabla \mathcal(J)$($\beta$) $\leftarrow$ $\frac{1}{Bn} \sum^B_{i=1} \sum^n_{j=1}[(-R^j_i$
        %        $\qquad \qquad$ $-b_i)\nabla_{\beta(\theta)}\log Prob_{\beta(\theta)}(\pi^j_i|s_i,\theta,\mathcal{F}^j_i)]$
        % \STATE $\beta$ $\leftarrow$ {\bfseries  Adam}($\beta,\nabla \mathcal(J)(\beta)$)
       \STATE $\mathcal{F}^j_i$ $\leftarrow$ $\mathcal{F}^j_i$ $\cup$ $\{\boldsymbol{f}(\pi_{i,1}), \boldsymbol{f}(\pi_{i,2})\}$ \quad $\forall i,j$ 
       \STATE $\boldsymbol{Y}$ $\leftarrow$ $\boldsymbol{Y}$ $\cup$ $\{ \frac{1}{B} \sum^B_{i=1} \boldsymbol{f}(\pi_{i,1}), \sum^B_{i=1} \boldsymbol{f}(\pi_{i,2})\}$ \quad $\forall i$
       \ENDFOR
    \STATE $\boldsymbol{Y}$ $\leftarrow$ {\bfseries LSSA}($\boldsymbol{Y}$)
    
    \STATE {\bfseries Output:}  The final solution set $\boldsymbol{Y}$
    \end{algorithmic}
\end{algorithm}

\section{Instance Augmentation}
Instance augmentation exploits multiple efficient transformations for the original instance that
share the same optimal solution. Then, all transformed problems are solved and the best solution
among them are finally selected. According to POMO, a 2D coordinate $(x, y)$ has eight different
transformations, $\{(x, y),(y, x),(x, 1-y),(y, 1-x),(1-x, y),(1-y, x),(1-x, 1-y),(1-y, 1-x)\}$. An instance of Bi-CVRP has 8 transformations,
and an instance of $m$-objective TSP has $8^m$ transformations \cite{lin2022pareto} due to the full transformation
permutation of $m$ groups of 2-dimensional coordinates,.

\section{An Alternative Form of CDE}
\label{gaplalter}
\begin{figure}[!htbp]
			\centering

                \includegraphics[width=0.3\textwidth]{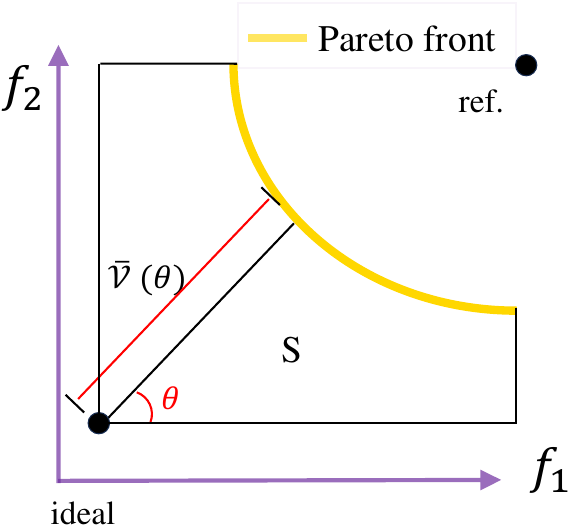}
                \caption{Alternative Pareto front hypervolume calculation in the polar coordinate.}
                \label{fig1a}
                % \vspace{-1.1cm}
\end{figure}

We also provide an alternative surrogate hypervolume function, denoted as CDE-alter. Figure \ref{fig1a} shows details. We  first
define region $S$ as the set of points dominating
the Pareto front.
\begin{eqnarray} 
            \begin{aligned}
            \label{alter1}
   &S=\{q|\exists p \in \mathcal{T}: p \prec q \quad and \quad q \succ p^{ideal}\}.
              \end{aligned}           
            \end{eqnarray}
We use the notation $\Lambda(\cdot)$ to represent the
Lebesgue measure of a set. Geometrically, as
illustrated in Figure \ref{fig1a}, it can be observed that:
\begin{eqnarray} 
            \begin{aligned}
            \label{alter3}
   &\Lambda(S) + \mathcal{HV}_r(Y)= \prod_{i=1}^m(r_i-z_i).
              \end{aligned}           
            \end{eqnarray}
The volume of $S$ can be calculated in a polar
coordinate as follows,
\begin{eqnarray} 
            \begin{aligned}
            \label{altert}
   &\Lambda(S) = \frac{\Phi}{m 2^m} \int_{(0,\frac{\pi}{2})^(m-1)} \bar{\mathcal{V}}(\theta).
              \end{aligned}           
            \end{eqnarray}

Thus, the Pareto hypervolume can be estimated as the volume difference between the regions
dominating $\boldsymbol{r}$ and those that dominate the Pareto front. CDE-alter maximizes the following objective,
\begin{eqnarray} 
            \begin{aligned}
            \label{alter}
   &\widetilde{\mathcal{HV}_r}(\beta, \boldsymbol{\theta}^P, s)=\prod_{i=1}^m(r_i-z_i)-
   \frac{\Phi}{m 2^m} \mathbb{E}_{ \boldsymbol{\theta}^P}[(\bar{\mathcal{V}}_{\beta}(\theta^i,s))^m]\\
   &\bar{\mathcal{V}}_{\beta}(\theta,s) = \max(\bar{\mathcal{G}}^{mtch}(f(\Psi_{\beta}(\theta, s)),\theta),0))
              \end{aligned},           
            \end{eqnarray}
where $\bar{\mathcal{G}}^{mtch}(\cdot)$  is an alternative projected distance function:
\begin{eqnarray} 
            \begin{aligned}
            \label{alter2}
   &\bar{\mathcal{G}}^{mtch}(f(\Psi_{\beta}(\theta, s)),\theta)=\max_{i \in \{1,...,m\}}\{(y_i-z_i)/\lambda_i(\theta)\}.
              \end{aligned}           
            \end{eqnarray}
 % of the calculation of Eq. \ref{alter} and Eq. \ref{alter1}.
Although the Pareto neural model $\Psi_{\beta}(\cdot)$ theoretically
has the ability to represent the complete Pareto set \cite{hornik1989multilayer}, empirical results show that the quality of the learned solutions is sensitive to the specific choice: Eq. \ref{comput} and Eq. \ref{alter} give different performances on various tasks. The main results of CDE-alter can be found in Appendix \ref{mmain}.

\section{Additional experiments}
\subsection{Main results of CDE-alter and Analysis}
\label{mmain}
As introduced in Appendix \ref{gaplalter}, we provide an alternative form of CDE, namely CDE-alter. Figure \ref{t22} shows the results of CDE-alter compared with several baselines. CDE-alter also shows the best overall performance. Here, we want to discuss the nature of CDE and CDE-alter since they show different features on various problems. Specifically, CDE-alter can show better performance on concave Pareto front (i.e. MOKP), while CDE-alter can show better performance on non-concave Pareto front (i.e. MOTSP and MOCVRP). In terms of non-concave Pareto front, the projection distance $\mathcal{V}_{\beta}(\theta)$ is uniformly distributed, which provides robustness for training model, while $\bar{\mathcal{V}}_{\beta}(\theta)$  is sensitive because of its non-uniform distribution which may lead to the overfitting of Pareto attention model. This phenomenon is the opposite on the question concave Pareto front. Therefore, it is necessary to choose different alternatives for specific problems.

\begin{table*}[!ht]
\caption{Experimental results on 200 random instances for MOCO problems.}
\centering
\scriptsize
\label{t22}
\begin{tabular}{l|ccc|ccc|ccc}
% \centering

\toprule
                       & \multicolumn{3}{c|}{Bi-TSP20}               & \multicolumn{3}{c|}{Bi-TSP50}               & \multicolumn{3}{c}{Bi-TSP100} \\
Method                   & HV $\uparrow$        & Gap $\downarrow$& Time $\downarrow$ & HV $\uparrow$       & Gap $\downarrow$ & Time $\downarrow$ & HV $\uparrow$      & Gap $\downarrow$  & Time $\downarrow$  \\ \midrule
WS-LKH (101 pref.)       &  0.6268         & 0.70\%   & 4.2m    &    0.6401        & 0.47\%   & 41m    &  \textbf{0.7071}         &\textbf{-0.06\%}   & 2.6h       \\
PPLS/D-C (200 iter.)     &  0.6256         & 0.89\%    &  25m    &   0.6282        & 2.32\%    &  2.7h    &  0.6845            & 3.14\%      &  11h      \\
NSGAII-TSP               &  0.5941        & 5.88\%    & 40m    &    0.5984       & 6.95\%    &  43m    &    0.6643          &  6.00\%     &    53m    \\ \midrule
DRL-MOA (101 models)     &  0.6257         &  0.87\%   & 7s   &    0.6360       & 1.10\%    & 10s    &      0.6971        & 1.36\%      &  22s      \\
PMOCO (101 pref.)       &   0.6266        & 0.73\%    &  8s   &  0.6349         &  1.28\%   & 13s   &    0.6953          & 1.61\%      &   21s     \\
NHDE-P (101 pref.)      &   0.6288        & 0.38\%    & 4.3m   &   0.6389        & 0.65\%    &  8.3m  &   0.7005           & 0.88\%      &    16m    \\
CDE-alter (101 pref.)      &  \underline{0.6301}         & \underline{0.17\%}    & 18s    &  0.6394        & 0.58\%    &  23s  &      0.7009        &  0.82\%     &  31s      \\ \midrule
EMNH (aug.) &  0.6271         & 0.65\%    & 1.3m   &  0.6408        & 0.36\%    &  4.6m   &    0.7023           & 0.62\%      &   17m \\
PMOCO (101 pref. \& aug.)   &  0.6273         & 0.62\%    & 46s   &   0.6392        & 0.61\%    &  3.1m    &   0.6997           & 0.99\%      &   5.7m     \\
NHDE-P (101 pref. \& aug.)   &  0.6296         &  0.25\%   & 9.8m   &  \underline{0.6429}         &\underline{0.03\%}     &  19m  &     0.7050         &  0.24\%     &   40m     \\
CDE-alter (101 pref. \& aug.) &  \textbf{0.6312}         & \textbf{0.00\%}    & 1.2m   &  \textbf{0.6431}        & \textbf{0.00\%}    &  4.5m   &   \underline{0.7067}           &  \underline{0.00\%}     & 6.8m       \\ \midrule     
Method                   & \multicolumn{3}{c|}{Bi-CVRP20}               & \multicolumn{3}{c|}{Bi-CVRP50}               & \multicolumn{3}{c}{Bi-CVRP100}  \\ \midrule
PPLS/D-C (200 iter.)     &  0.3351         & 3.98\%    &  1.2h    &  0.4149         & 3.31\%    &  9.6h    &  0.4083            &  1.80\%     &  37h      \\
NSGAII-CVRP               &   0.3123        & 10.52\%    & 37m    &  0.3631         &  15.38\%   &  38m    &  0.3538            &  14.91\%     &  43m      \\ \midrule

DRL-MOA (101 models)     &     0.3453     & 1.06\%    & 7s   &  0.4270         &  0.49\%   &  20s   &  \textbf{0.4176}          &   \textbf{-0.43\%}    &  40s      \\
PMOCO (101 pref.)       &   0.3467        & 0.66\%    &  8s    &   0.4271        & 0.47\%    &  18s  &   0.4131           &  0.65\%     &   36s     \\
NHDE-P (101 pref.)      &  0.3458        & 0.92\%    & 1.5m   &   0.4248        & 1.00\%    & 3.1m   &  0.4127            &  0.75\%     &   5.3m     \\
CDE-alter (101 pref.)      &  0.3468         & 0.63\%    & 17s   &   0.4272        & 0.44\%    & 31s  &     0.4140        &  0.43\%     &  58s      \\ \midrule
EMNH (aug.) &  0.3471         & 0.54\%    & 33s  &  0.4250        & 0.96\%    &  1.4m  &    0.4146           & 0.29\%      &   3.7m \\
PMOCO (101 pref. \& aug.)   & \underline{0.3481}          & \underline{0.26\%}    & 1m   &   \underline{0.4287}        & \underline{0.09\%}    &  2.1m    &   0.4150           &  0.19\%     &    4.5m    \\
NHDE-P (101 pref. \& aug.)   &    0.3465      &  0.72\%  &  5.1m   &    0.4262       & 0.68\%    & 9.2m   &  0.4139            & 0.46\%      & 21m       \\
CDE-alter (101 pref. \& aug.) &    \textbf{0.3490}       &  \textbf{0.00\%}   & 2.2m   &    \textbf{0.4291}       & \textbf{0.00\%}    & 4.1m    &   \underline{0.4158}        &   \underline{0.00\%}    &   6.8m     \\ \midrule  
Method                   & \multicolumn{3}{c|}{Bi-KP50}               & \multicolumn{3}{c|}{Bi-KP100}               & \multicolumn{3}{c}{Bi-KP200}  \\ \midrule
WS-DP (101 pref.)       &  0.3563         & 0.83\%   & 9.5m    &    0.4531        & 1.03\%   & 1.2h   &  0.3599         & 2.20\%    & 3.7h       \\
PPLS/D-C (200 iter.)     &  0.3528         & 1.81 \%    &  18m    &   0.4480        &  2.14\%   & 46m     &  0.3541        & 3.78\%      &  1.4h      \\
NSGAII-KP               &  0.3112         &  13.39\%   &  30m   &    0.3514       &  23.24\%   & 31m     &  0.3511       &  4.59\%     &    33m    \\ \midrule
EMNH              &  0.3561        & 0.89\%    & 17s  &  0.4535        & 0.94\%    &  53s  &   0.3603          & 2.09\%      &   2.3m\\
DRL-MOA (101 models)     &  0.3559         & 0.95\%    &  8s  &  0.4531         & 1.03\%    &  13s   &    0.3601          &  2.15\%     &  1.1m      \\
PMOCO (101 pref.)       &   0.3552        &  1.14\%   &  13s    &  0.4522         & 1.22\%    & 19s   &   0.3595           & 2.31\%      &   50s     \\
NHDE-P (101 pref.)      &   \underline{0.3564}        & \underline{0.81\%}    & 1.1m   &   \underline{0.4541}        & \underline{0.81\%}    &  2.5m  &   \underline{0.3612}           & \underline{1.85\%}      &  5.3m      \\
CDE-alter (101 pref.)      &   \textbf{0.3593}        & \textbf{0.00\%}    &  21s  &   \textbf{0.4578}        &  \textbf{0.00\%}   &  33s &    \textbf{0.3680}          &   \textbf{0.00\%}    &   1.4m     \\ \midrule  
Method                   & \multicolumn{3}{c|}{Tri-TSP20}               & \multicolumn{3}{c|}{Tri-TSP50}               & \multicolumn{3}{c}{Tri-TSP100}  \\ \midrule
WS-LKH (210 pref.)       &  0.4718         &  1.34\%   & 20m    &    0.4493        &  2.30\%   & 3.3h   &   0.5160      & 1.02\%    & 11h       \\
PPLS/D-C (200 iter.)     &  0.4698         & 1.76\%    & 1.3h     &   0.4174        & 9.24\%    &  3.8h    &    0.4376          & 16.06\%      & 13h       \\
NSGAII-TSP               &  0.4216         & 11.84\%    & 2.1h    &   0.4130        &  10.20\%   &  2.3h    &  0.4291            &  17.69\%     &  2.5h      \\ \midrule
DRL-MOA (1035 models)     &   0.4712        & 1.46\%    & 51s   &  0.4396          & 4.41\%    & 1.5s    &    0.4915           & 5.72\%      &  3.1s      \\
PMOCO (10201 pref.)       &  0.4749         & 0.69\%    &  8.9m    &  0.4489         & 2.39\%    & 17m   &    0.5102          &  2.13\%     &  34m      \\
NHDE-P (10201 pref.)      &  0.4764         & 0.38\%    &  53m  &  0.4513         &  1.87\%   & 1.8h   &  0.5118            &  1.82\%     &  4.3h      \\
CDE-alter (10201 pref.)      &  0.4773         & 0.19\%    & 10m   &   0.4517        & 1.78\%    & 19m  &  0.5121            &  1.76\%     &    41m    \\ \midrule
EMNH (aug.)             &  0.4712        & 1.46\%    & 7.1m  &  0.4418        & 3.94\%    &  58m  &   0.4973          & 4.60\%      &   2.4h\\
PMOCO (10201 pref. \& aug.)   &  0.4757          & 0.52\%    & 20m   &   0.4573        & 0.57\%    &   1.1h   &    0.5123          & 1.73\%      &  4.3h      \\
NHDE-P (10201 pref. \& aug.)   &  \underline{0.4772}         & \underline{0.21\%}    &  2.1h  &   \underline{0.4595}        & \underline{0.09\%}    &  6.7h  &     \underline{0.5210}         &  \underline{0.06\%}     &   15.3h     \\
CDE-alter (10201 pref. \& aug.) &   \textbf{0.4782}        & \textbf{0.00\%}    & 26m   &  \textbf{0.4599}         &  \textbf{0.00\%}   &  1.3h   &    \textbf{0.5213}          &  \textbf{0.00\%}     &  4.8h      \\ \midrule
\end{tabular}
% \vspace{-0.7cm}
\end{table*}

\subsection{Generalization Study of CDE and CDE-alter}
To assess the generalization capability of CDE and CDE-alter. We compare them with the other baselines on 200 random Bi-TSP instances with larger sizes, i.e., Bi-TSP150/200. The comparison results are demonstrated in Table \ref{t222}. CDE and CDE-alter outperform
the state-of-the-art MOEA (i.e., PPLS/D-C) and other neural methods significantly, in terms of
HV, which means a superior generalization capability.  Note that WS-LCK has a slight advantage over  CDE and CDE-alter at the sacrifice of Dozens of times the inference time (i.e. 1.2 hours vs 4.6 minutes). Besides, CDE is superior to CDE-alter on both problems, which also Verifies our idea in Appendix \ref{mmain}.
\begin{table*}[!ht]
\caption{Experimental results on 200 random instances for MOCO problems.}
\centering
\scriptsize
\label{t222}
\begin{tabular}{l|l|ccc|ccc}
% \centering

\toprule
           &            & \multicolumn{3}{c|}{Bi-TSP150}               & \multicolumn{3}{c}{Bi-TSP200}               \\
Type&Method                   & HV $\uparrow$        & Gap $\downarrow$& Time $\downarrow$ & HV $\uparrow$       & Gap $\downarrow$ & Time $\downarrow$       \\ \midrule
\multicolumn{1}{l|}{\multirow{3}{*}{Traditional heuristics}}
    &WS-LKH (101 pref.)       &  \textbf{0.7075}         & \textbf{-0.77\%}   & 6.4h       &   \textbf{0.7435}        & \textbf{-1.31\%}   &  13h          \\
&PPLS/D-C (200 iter.)     &  0.6784         & 3.38\%    &  21h    &   0.7106         & 3.17\%    &  32h          \\
&NSGAII-TSP               &  0.6125        & 12.76\%    & 2.1h    &    0.6318       & 13.91\%    &  3.7h       \\ \midrule
\multicolumn{1}{l|}{\multirow{5}{*}{Neural heuristics}}
&DRL-MOA (101 models)     &  0.6901         &  1.71\%   & 45s   &    0.7219       & 1.64\%    & 2.1m         \\
&PMOCO (101 pref.)       &   0.6912        & 1.55\%    &  1.4m  &  0.7231         &  1.47\%   & 3.1m        \\
&NHDE-P (101 pref.)      &    0.6964        & 0.81\%    & 12m   &    0.7280        & 0.80\%    &  21m         \\
&CDE-alter (101 pref.)      &  0.6953         & 0.97\%    & 2.3m    &  0.7273        & 0.90\%    &  4.6m     \\ 
&CDE (101 pref.)      &  0.6972         & 0.70\%    & 2.3m    &  0.7291        & 0.65\%    &  4.6m     \\ 
\midrule
\multicolumn{1}{l|}{\multirow{5}{*}{\makecell{Neural heuristics\\augmentation}}}
&EMNH (aug.) &  0.6983         & 0.54\%    & 53m   &  0.7307        & 0.44\%    &  2.9h   \\
&PMOCO (101 pref. \& aug.)   &  0.6967         & 0.77\%    & 41m   &   0.7276        & 0.86\%    &  2.9h         \\
&NHDE-P (101 pref. \& aug.)   &  0.7012         &  0.13\%   & 1.3m   &  0.7324         &0.20\%     &  4.6h     \\
&CDE-alter (101 pref. \& aug.) &  0.7003         & 0.26\%    & 53m   & 0.7316        & 0.31\%    &  1.2h        \\
&CDE (101 pref. \& aug.) &  \underline{0.7021}         & \underline{0.00\%}    & 53m   &  \underline{0.7339}        & \underline{0.00\%}    &  1.2h\\
\midrule   \end{tabular}
% \vspace{-0.7cm}
\end{table*}

\subsection{More Hyperparameter Studies}
\label{Hyperparameter}
We further study the effects of $P'$ (maximal size of polar
angles pool) and $c$ (control parameter of potential energy).

\begin{figure}
    \centering
    % \begin{subfigure}[b]{0.49\linewidth}
    %     \centering
    %     \includegraphics[width=\linewidth]{dtlzd6.pdf}
    % \end{subfigure}
    % 238 227
    \subfigure[Bi-TSP($P'$)]{\begin{minipage}[h]{0.23\textwidth}
        \includegraphics[width=1\textwidth]{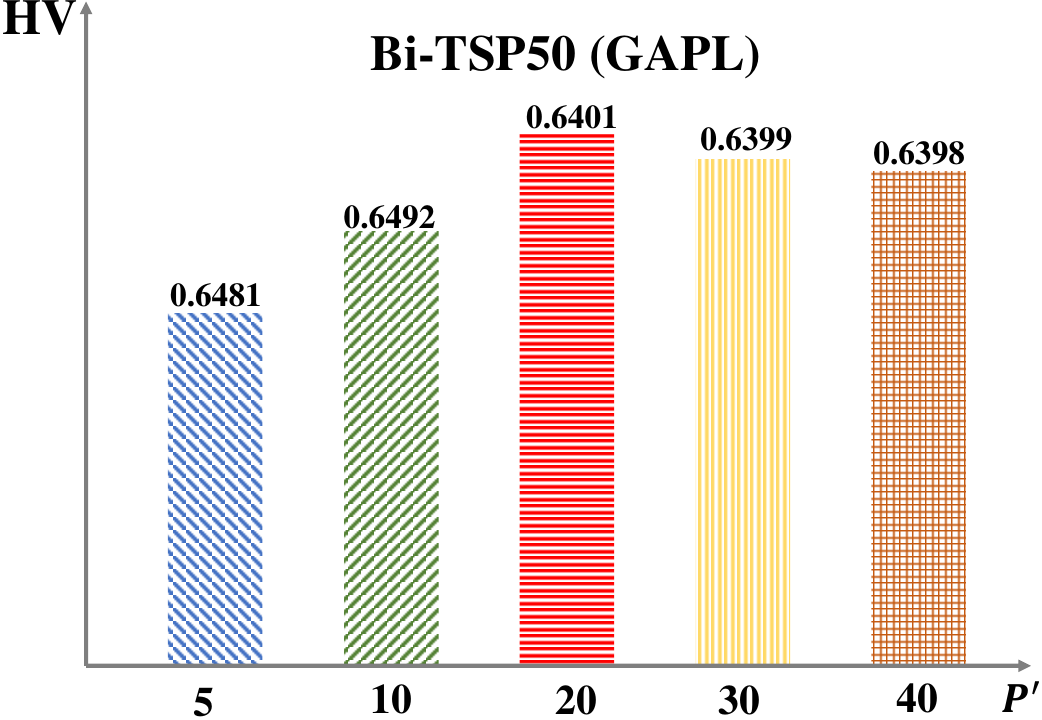}
    \end{minipage}}
    \subfigure[Tri-TSP($P'$)]{\begin{minipage}[h]{0.23\textwidth}
        \includegraphics[width=1\textwidth]{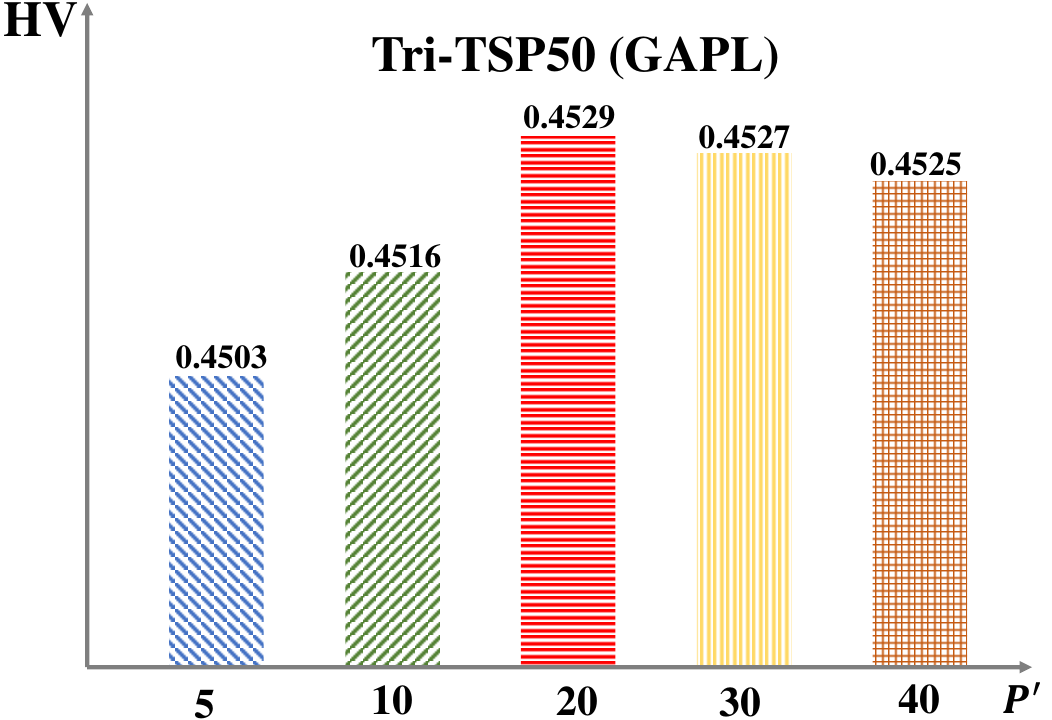}
    \end{minipage}}
        \subfigure[Bi-TSP($c$)]{\begin{minipage}[h]{0.23\textwidth}
    % \vspace{-1cm}
        \includegraphics[width=1\textwidth]{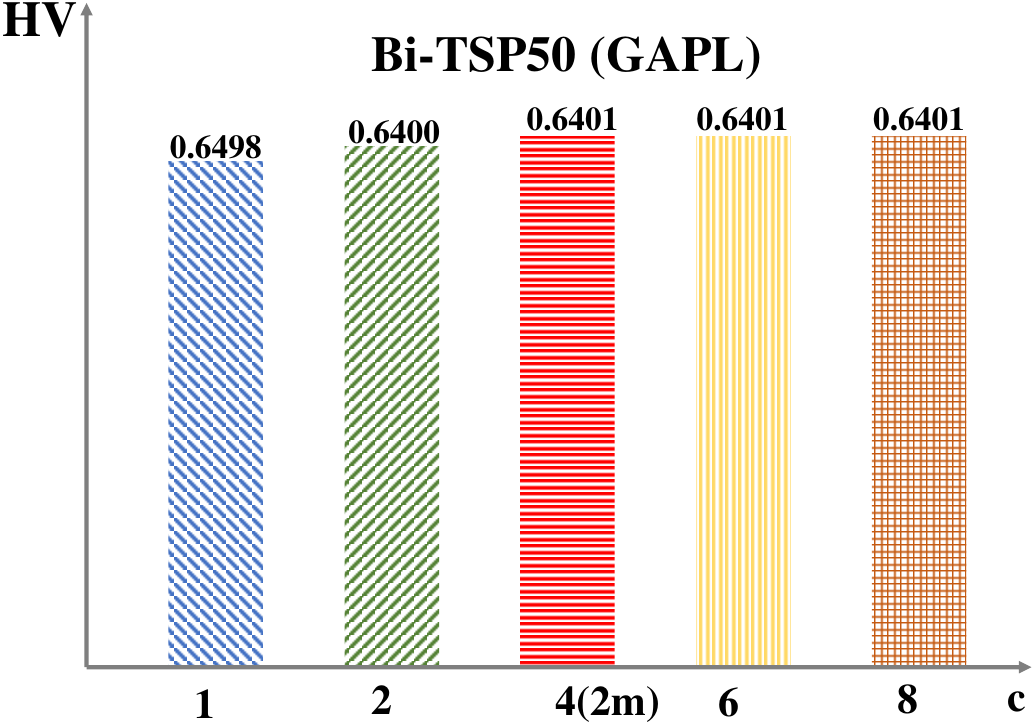}
    \end{minipage}}
    \subfigure[Tri-TSP($c$)]{\begin{minipage}[h]{0.23\textwidth}
    % \vspace{-1cm}
        \includegraphics[width=1\textwidth]{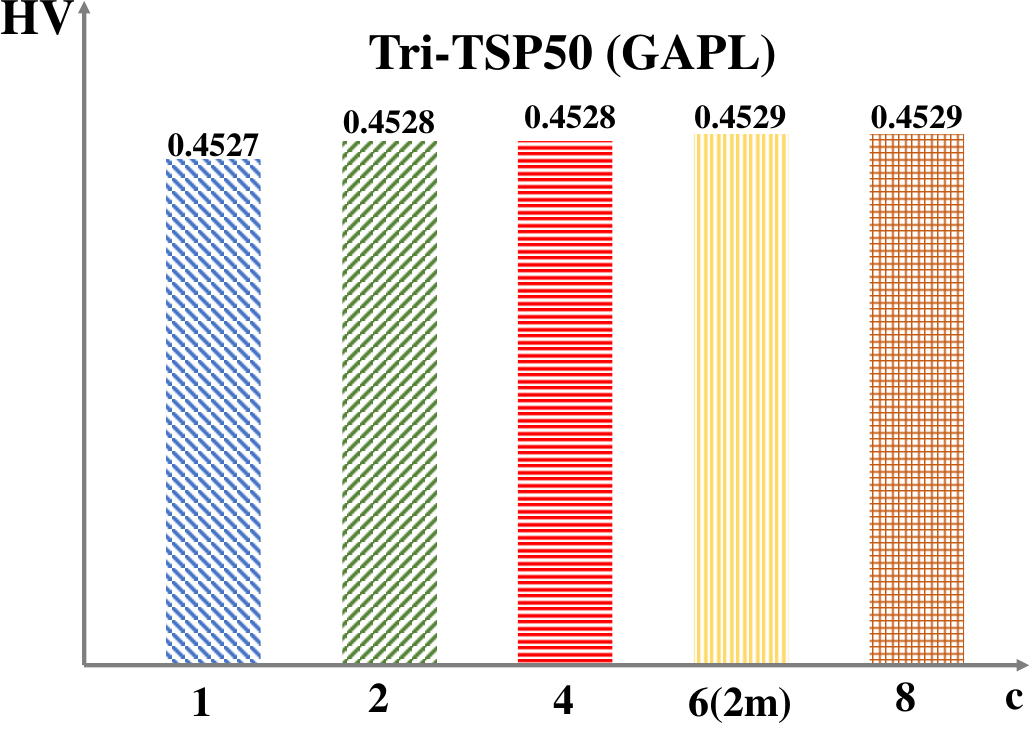}
    \end{minipage}}
    \caption{Hyperparameter study. (a) Effect of the maximal size of polar angles pool used in training on Bi-TSP50. (b) Effect of the maximal size of polar angles pool used in training on Tri-TSP100. (c) Effect of the control parameter of potential energy on Bi-TSP50.  (d) Effect of the control parameter of potential energy on Tri-TSP50.
points from new solutions for updating the Pareto front.}
    \label{hy}
\end{figure}

We present the results of various $P'$ on Bi-TSP50 and Tri-TSP100 in Figure \ref{hy} (a) (b). As shown, $P'=5$ and $P'=10$ cause inferior performance, while proper $P'$ (20 $\leq$ 
 $P'$ $\leq$ 40) results in desirable performance. Intuitively, when limiting the same total gradient steps in training,  a larger $P'$ means fewer instances are used for model training. In this sense, too large $P'$, i.e., with insufficient instances, could lead to inferior performance for solving unseen instances. On the other hand, although the approximate HV is an unbiased estimation, when $P'$ is too small, the calculated variance will become larger, which will also prevent the model from learning favorable global information and thus deteriorate the final
performance. Hence we choose $P'=20$  in this paper.

Figure \ref{hy} (c) (d) displays the results of various values of $c$. As has been presented in Section \ref{novel}, 
the potential energy function is as Eq. \ref{Energya}, 
$c$ is a parameter that influences the result former literature \cite{blank2020generating} has reported that $c$ is a weak parameter, because the result does not change a lot when $c$ varies, and the
selection of this parameter is only related to the dimension of
the objective space. In this part, an experiment is conducted to
study the influence of $c$ on Bi-TSP50 and Tri-TSP100 in Figure \ref{hy} (c) (d), the results also support the proposition
of former researches. So, in this paper, we recommend to use
$c$ = $2m$ in LSS. 

\begin{figure}[!t]
\centering
    \subfigure[TCH(Bi-TSP)]{\begin{minipage}[h]{0.23\textwidth}
        \includegraphics[width=1\textwidth]{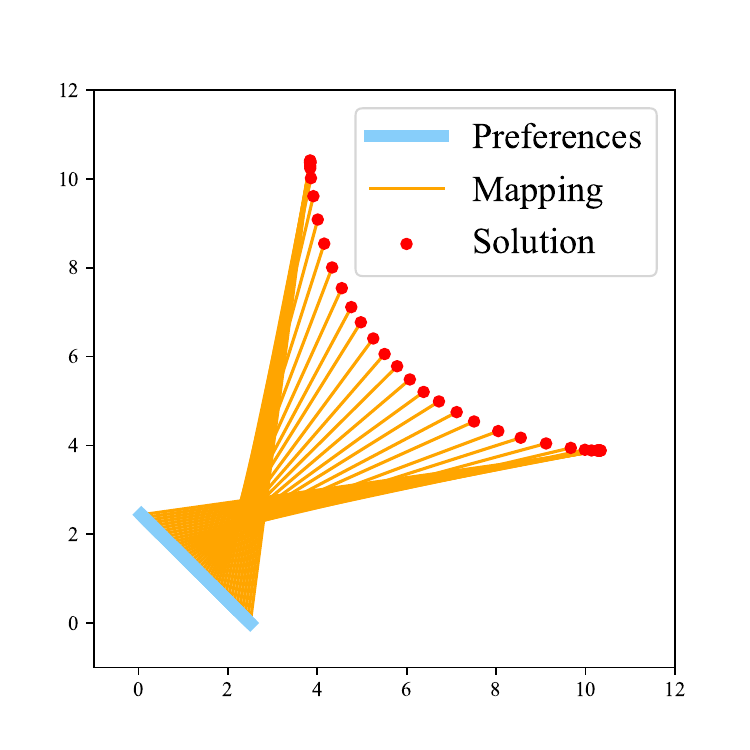}
    \end{minipage}}
    \subfigure[WS]{\begin{minipage}[h]{0.23\textwidth}
        \includegraphics[width=1\textwidth]{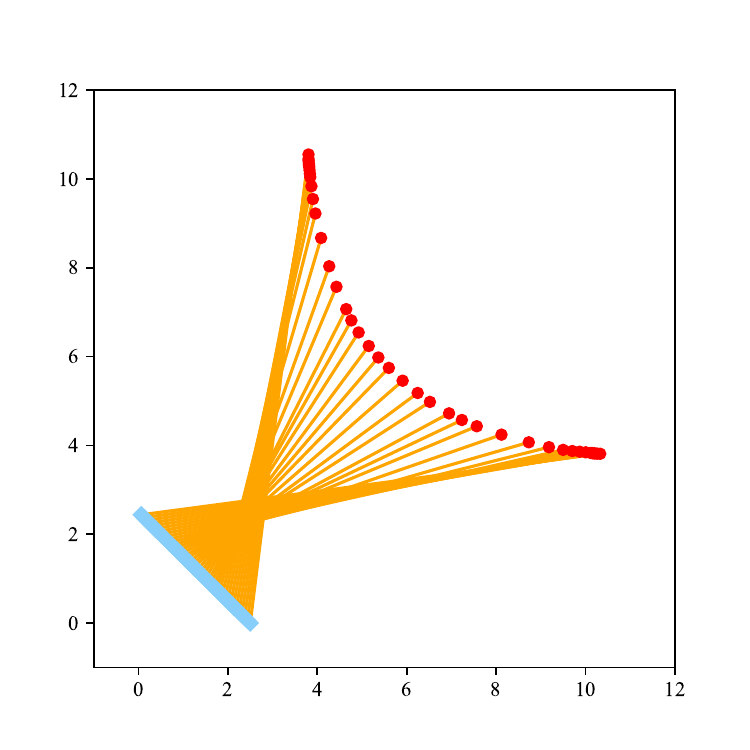}
    \end{minipage}}
    \subfigure[CDE]{\begin{minipage}[h]{0.23\textwidth}
    % \vspace{-1cm}
        \includegraphics[width=1\textwidth]{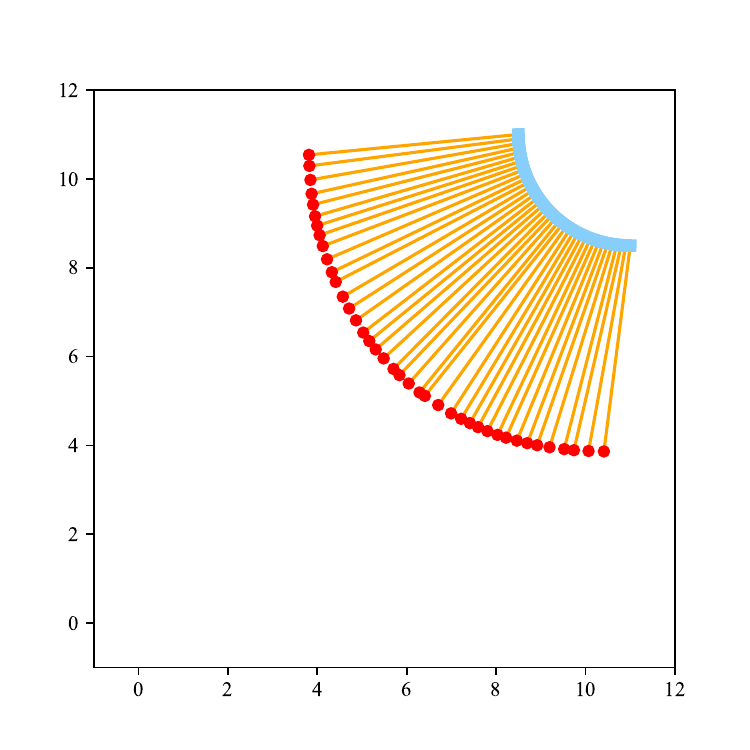}
    \end{minipage}}
    \subfigure[CDE-alter]{\begin{minipage}[h]{0.23\textwidth}
    % \vspace{-1cm}
        \includegraphics[width=1\textwidth]{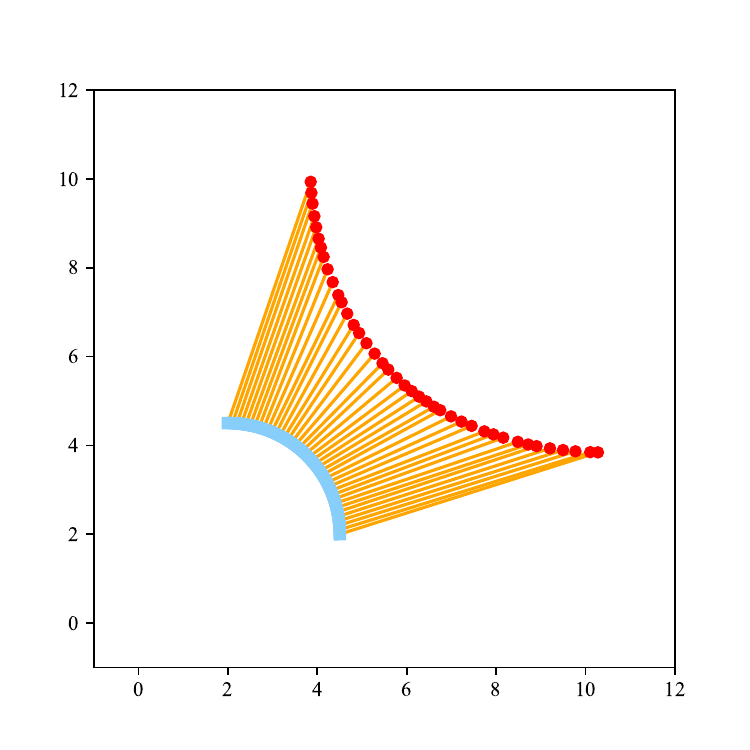}
    \end{minipage}}
        \subfigure[TCH(Bi-CVRP)]{\begin{minipage}[h]{0.23\textwidth}
        \includegraphics[width=1\textwidth]{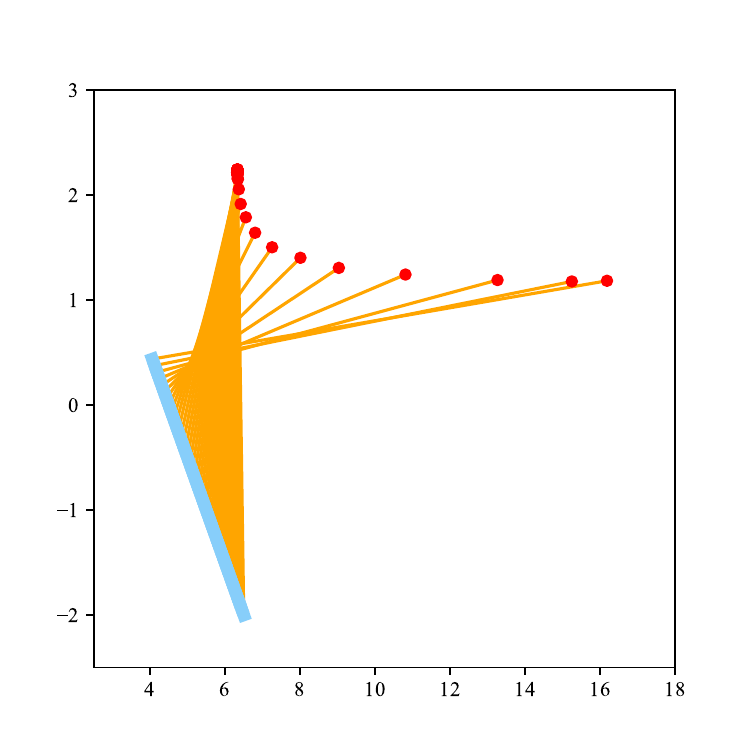}
    \end{minipage}}
    \subfigure[WS]{\begin{minipage}[h]{0.23\textwidth}
        \includegraphics[width=1\textwidth]{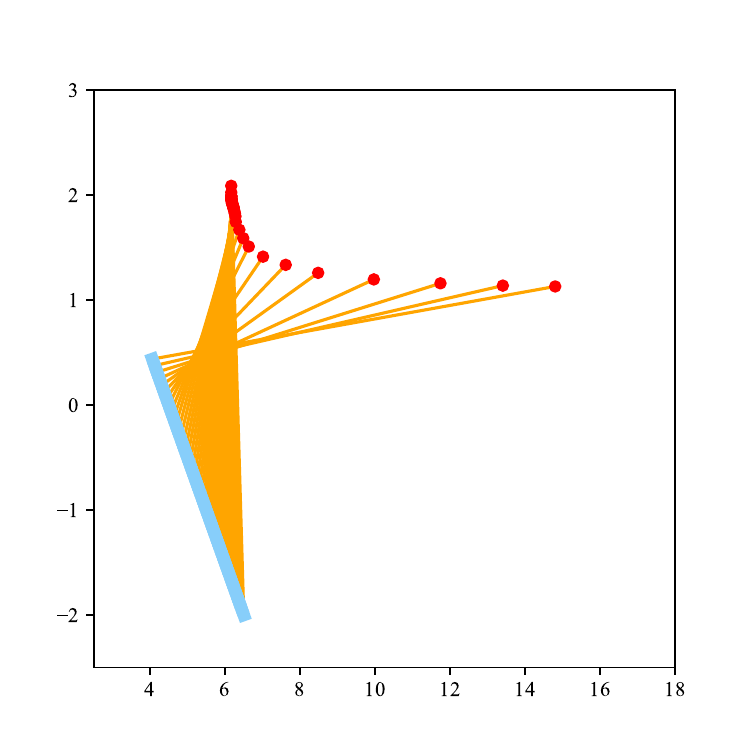}
    \end{minipage}}
        \subfigure[CDE]{\begin{minipage}[h]{0.23\textwidth}
        \includegraphics[width=1\textwidth]{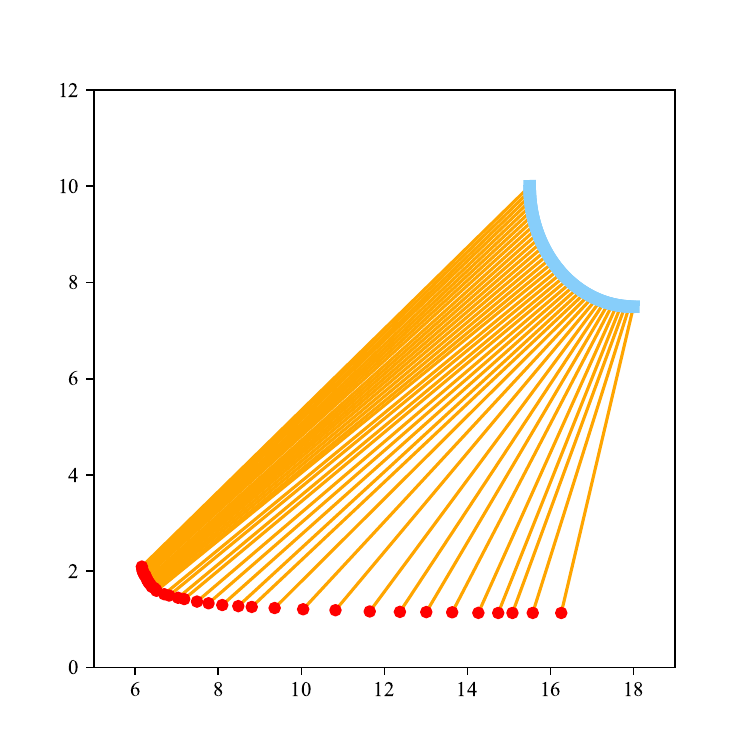}
    \end{minipage}}
    \subfigure[CDE-alter]{\begin{minipage}[h]{0.24\textwidth}
    % \vspace{-1cm}
        \includegraphics[width=1\textwidth]{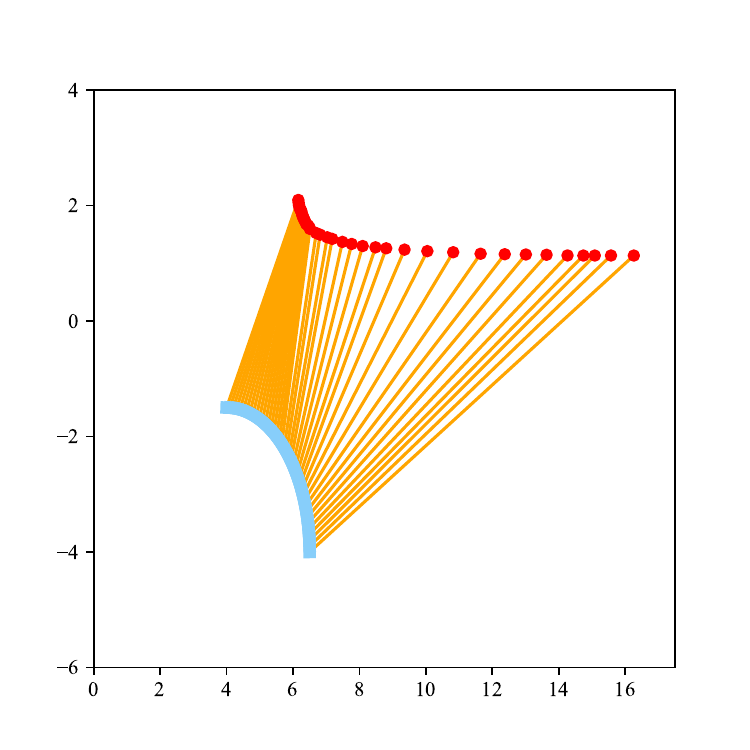}
    \end{minipage}}
    \subfigure[TCH(Bi-KP)]{\begin{minipage}[h]{0.23\textwidth}
        \includegraphics[width=1\textwidth]{TCHKP.pdf}
    \end{minipage}}
    \subfigure[WS]{\begin{minipage}[h]{0.23\textwidth}
        \includegraphics[width=1\textwidth]{WSKP.pdf}
    \end{minipage}}
        \subfigure[CDE]{\begin{minipage}[h]{0.23\textwidth}
        \includegraphics[width=1\textwidth]{GADKP.pdf}
    \end{minipage}}
    \subfigure[CDE-alter]{\begin{minipage}[h]{0.23\textwidth}
    % \vspace{-1cm}
        \includegraphics[width=1\textwidth]{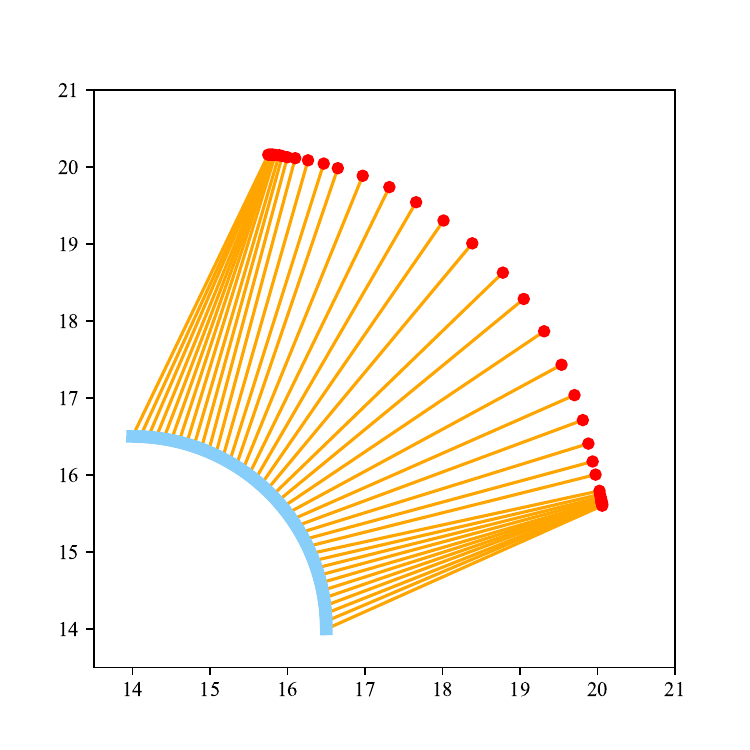}
    \end{minipage}}

    \subfigure[TCH(Tri-TSP)]{\begin{minipage}[h]{0.23\textwidth}
        \includegraphics[width=1\textwidth]{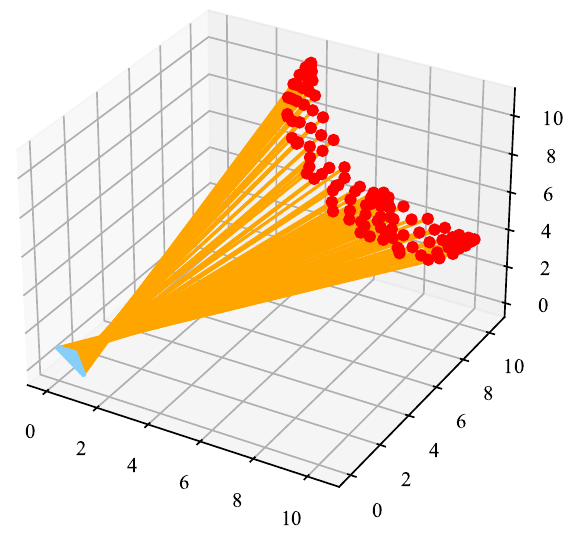}
    \end{minipage}}
    \subfigure[WS]{\begin{minipage}[h]{0.23\textwidth}
        \includegraphics[width=1\textwidth]{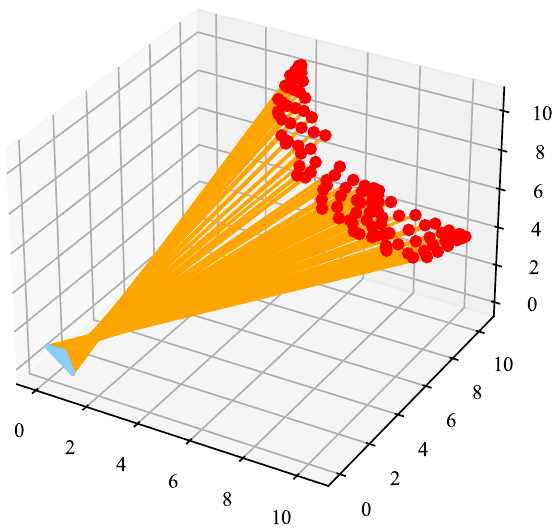}
    \end{minipage}}
        \subfigure[CDE]{\begin{minipage}[h]{0.23\textwidth}
        \includegraphics[width=1\textwidth]{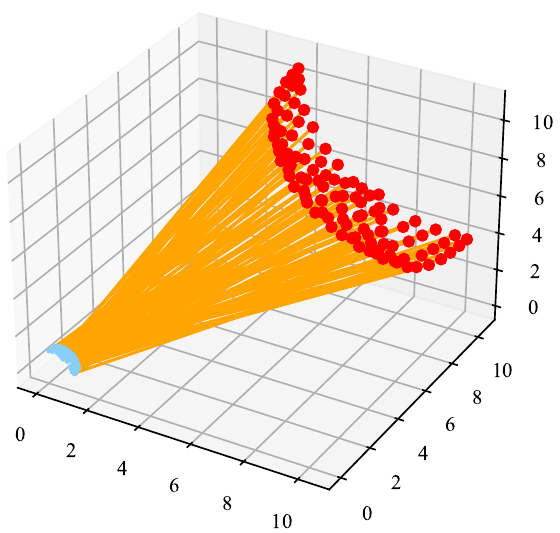}
    \end{minipage}}
    \subfigure[CDE-alter]{\begin{minipage}[h]{0.23\textwidth}
    % \vspace{-1cm}
        \includegraphics[width=1\textwidth]{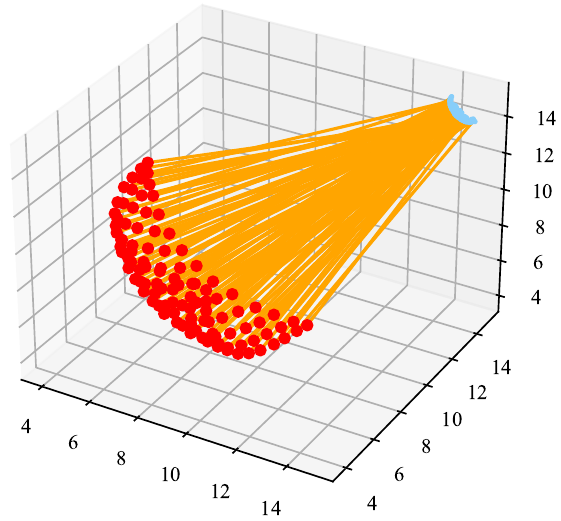}
    \end{minipage}}
    \caption{Visual comparisons on Bi-TSP20, Bi-CVRP, Bi-KP20 and Tri-TSP20. CDE  and GADL-alter have better adaptability than WS- and TCH-based approaches.}
    \label{mse1}
\end{figure}

\subsection{More  Visualization Results for Validity of Context Awareness}
\label{more}
We have shown  visualization Results of all three classic
MOCO problems for each decomposition method in Figure \ref{mse1}. All results can support the conclusion in Section \ref{validity} to reflect the superiority of CDE's context awareness.

\subsection{Effectiveness of LSSA}
\begin{figure}[H]
			\centering

                \includegraphics[width=0.4\textwidth]{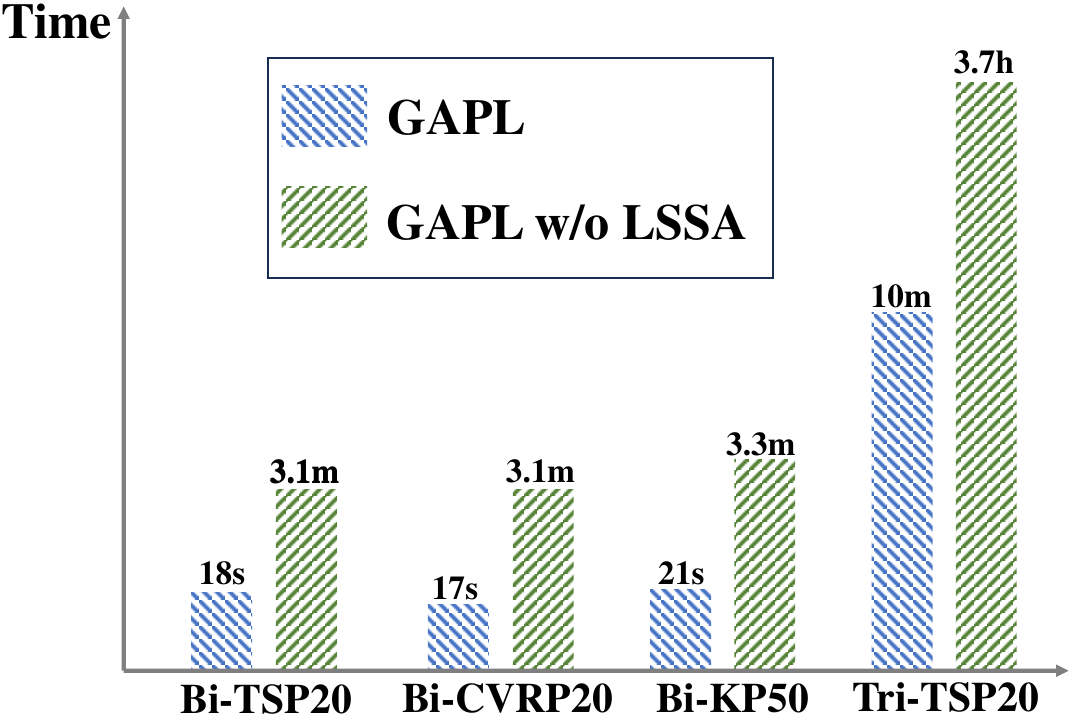}
                \caption{The time consumed by CDE and CDE w/o LSSA.} 
                % on 4 problems
                % strategy improves efficient to a great extent.}
                \label{time}

\end{figure}

We also show the inference time of CDE and CDE w/o LSSA in Figure \ref{time}. CDE w/o LSSA adopts traditional LSS. It is evident that LSSA 
can improve efficiency, as discussed in Section \ref{44}.

\section{Convergence of the Pareto Front Hypervolume}
Assuming that $b \leq r_1-y_i \leq B, \forall i \in \{1,...,m\}$, $y \in Y$.  Let $\mathcal{Z}(\theta)= \frac{\Phi}{m2^m}\mathcal{V}_{\mathcal{X}}(\theta)^m$, then $sup_{\theta}\mathcal{Z}(\theta) \leq \frac{\Phi}{m2^m} B^m m^{m/2}$. Let $\hat{\mathcal{HV}}_{\boldsymbol{r}}(Y) := \frac{1}{P}\sum_{i=1}^P\mathcal{Z}(\theta^{(i)})$ denote the empirical estimation of ${\mathcal{HV}}_{\boldsymbol{r}}(Y)$ with $P$ samples. Via Hoeffding inequality, similar to \cite{zhang2020random}, we have the following inequality:
\begin{eqnarray} 
            \begin{aligned}
            \label{alter222}
   &\rm{Pr}(|\hat{\mathcal{HV}}_{\boldsymbol{r}}(Y)-{\mathcal{HV}}_{\boldsymbol{r}}(Y)| \geq \epsilon) \leq 2 exp(\frac{-P\epsilon^2  2^{m+2}}{\Phi^2 B^{2m}m^{m-2}}).
              \end{aligned}           
            \end{eqnarray}

\section{The bound of $\mathcal{V}_{\mathcal{X}}(\theta)$ in Eq. \ref{comput}}
\label{proof2}
Since $\mathcal{V}_{\mathcal{X}}(\theta)$ is a max-min problem, we can conclude that the following inequalities hold:
\begin{eqnarray} 
            \begin{aligned}
            \label{alter444444}
   \mathcal{V}_{\mathcal{X}}(\theta) \leq Bm^{\frac{1}{2}},
              \end{aligned}           
            \end{eqnarray}
where $b \leq (r_i-f_i(\boldsymbol{x})) \leq B, \forall \boldsymbol{x} \in \mathcal{X}, \forall i \in \{1,...,m\}$ and $\|\boldsymbol{\lambda}\|_2=1$. 

$Proof.$ 
\begin{eqnarray} 
            \begin{aligned}
            \label{alter333}
   \mathcal{V}_{\mathcal{X}}(\theta) &\leq \max_{\boldsymbol{x} \in \mathcal{X}, \|\boldsymbol{\lambda}\|_2=1} \left(\min_{i \in \{1,...,m\}}\left\{\frac{r_i-f_i(\boldsymbol{x})}{\lambda_i(\theta)}\right\}\right) \\
    &(r_i-f_i(\boldsymbol{x}) \leq B) \\
    &\leq \max_{\|\boldsymbol{\lambda}\|_2=1}\left(\min_{i \in \{1,...,m\}}\left\{\frac{B}{\lambda_i(\theta)}\right\}\right) \\
    &\leq \frac{B}{m^{-\frac{1}{2}}}\\
    &=Bm^{\frac{1}{2}}.
              \end{aligned}           
            \end{eqnarray}
The transition from line one to line two is due to the fact that the inequality  $r_i-f_i(\boldsymbol{x}) \leq B$  holds for $\forall \boldsymbol{x} \in \mathcal{X}$   and for  $\forall i \in \{1,...,m\}$. The transition from line two to line three is   $\max_{\|\boldsymbol{\lambda}\|_2=1}\left(\min_{i \in \{1,...,m\}}\left\{\frac{B}{\lambda_i(\theta)}\right\}\right)$  is an optimization problem under the constraint $\|\boldsymbol{\lambda}\|_2=1$. The upper bound for this optimization is when $\lambda_1=...=\lambda_m=m^{-\frac{1}{2}}$. \hfill $\square$   
% \end{document}
\section{Equivalence of Hypervolume Calculation in Polar Coordinates}
\label{proof3}
\textit{Proof.} ${\mathcal{HV}}_{\boldsymbol{r}}(Y)$ can be simplified by the following equations. Here $\Omega$ denoted the dominated regions
by the Pareto front, i.e., ${\mathcal{HV}}_{\boldsymbol{r}}(Y)=\Lambda(\Omega,\textbf{r})$.

\begin{eqnarray}
            \begin{aligned}
            \label{alter111}
   {\mathcal{HV}}_{\boldsymbol{r}}(Y) &= \int_{\mathcal{R}^m} I_{\Omega}dy_1...y_m\\
        &(``dv''\enspace \rm{denoted} \enspace \rm{the} \enspace\rm{infinitesimal} \enspace \rm{sector} \enspace\rm{area.} )\\
             &=\underbrace{\int^{\frac{\pi}{2}}_0 ... \int^{\frac{\pi}{2}}_0}_{m-1}dv \\
             &(``d'' \enspace \rm{equals}\enspace \rm{the}\enspace \rm{angle}\enspace \rm{ratio}\enspace \rm{multiplied}\enspace \rm{by}\enspace \mathcal{V}_{\mathcal{X}}(\theta)^m\enspace \rm{multiplied}\enspace \rm{by}\enspace \rm{the}\enspace \rm{unit}\enspace \rm{for}\enspace \rm{volume.} ) \\
             &=\underbrace{\int^{\frac{\pi}{2}}_0 ... \int^{\frac{\pi}{2}}_0}_{m-1} \frac{\Phi}{m} \cdot \frac{\mathcal{V}_{\mathcal{X}}(\theta)^m}{2\pi \cdot \pi^{m-2}}\underbrace{d\theta_1...\theta_{m-1}}_{d\theta} \\
             &=\frac{\Phi}{2m\pi^{m-1}}\underbrace{\int^{\frac{\pi}{2}}_0 ... \int^{\frac{\pi}{2}}_0}_{m-1}\mathcal{V}_{\mathcal{X}}(\theta)^m d\theta \\
             &=\frac{\Phi}{2m\pi^{m-1}} \cdot {\frac{\pi}{2}}^{m-1} \cdot \mathbb{E}_{\theta}[\mathcal{V}_{\mathcal{X}}(\theta)^m]\\
             &=\frac{\Phi}{m2^{m}}[\mathcal{V}_{\mathcal{X}}(\theta)^m].
              \end{aligned}           
            \end{eqnarray}
We specify $\theta \sim \rm{Unif}(\Theta) = \rm{Unif}([0,\frac{\pi}{2}]^{m-1})$ in Eq. \ref{alter111}.

Line 2 holds since it represents the integral of $\Omega$ expressed in polar coordinates, where in the element $dv$  corresponds to the volume associated with a segment obtained by varying  $d\theta$.

Line 3 calculates the infinitesimal volume of $dv$ by noticing the fact that the ratio of  $dv$ to $\frac{\Phi}{m}$ is $\frac{\mathcal{V}_{\mathcal{X}}(\theta)^m}{2\pi \cdot \pi^{m-2}}$. Line 4 is a simplification of Line 3. And Line 5 and 6 express the integral in its expectation
form.  \hfill $\square$
% This document was modified from the file originally made available by
% Pat Langley and Andrea Danyluk for ICML-2K. This version was created
% by Iain Murray in 2018, and modified by Alexandre Bouchard in
% 2019 and 2021 and by Csaba Szepesvari, Gang Niu and Sivan Sabato in 2022.
% Modified again in 2023 and 2024 by Sivan Sabato and Jonathan Scarlett.
% Previous contributors include Dan Roy, Lise Getoor and Tobias
% Scheffer, which was slightly modified from the 2010 version by
% Thorsten Joachims & Johannes Fuernkranz, slightly modified from the
% 2009 version by Kiri Wagstaff and Sam Roweis's 2008 version, which is
% slightly modified from Prasad Tadepalli's 2007 version which is a
% lightly changed version of the previous year's version by Andrew
% Moore, which was in turn edited from those of Kristian Kersting and
% Codrina Lauth. Alex Smola contributed to the algorithmic style files.

\section{Licenses}
\label{lic}
The licenses for the codes  we used in this work are shown in Table \ref{licenses}.
\begin{table}[htbp]  
    %  \scriptsize
    
    \scriptsize
        % \tiny\scriptsize\footnotesize\small\normalsize\large\Large\LARGE\huge\Huge
 % \tabcolsep0.01in
 \renewcommand{\arraystretch}{1.3}
  \caption{List of licenses for the codes suite we used in this work.}
  \label{licenses}
    \centering
  \begin{tabular}{lllll}
    \toprule
    Resources&Type&Link&License\\
    \midrule
    PSL-MOCO \cite{lin2022pareto} &Code&https://github.com/Xi-L/PMOCO &MIT License\\
    NHDE \cite{chen2023neural} & Code&https://github.com/Bill-CJB/NHDE &MIT License\\
    EMNH \cite{chen2023efficient}&Code&https://github.com/Bill-CJB/EMNH& MIT License\\
    
    \bottomrule
    
  \end{tabular}
\end{table}

\end{document}